\documentclass{article}

\usepackage[english]{babel}

\usepackage[letterpaper,top=2cm,bottom=2cm,left=3cm,right=3cm,marginparwidth=1.75cm]{geometry}% arts, genealogy, histories, humanities, jintelligence, laws, literature, religions, risks, socsci
\usepackage{soul}
\usepackage{xcolor}  % For custom highlight colors
\usepackage{amsmath}
\usepackage{graphicx}
\usepackage{natbib}
\usepackage{float}
\usepackage{subfig}
\usepackage{url}
\sethlcolor{green}
\newcommand{\corr}[1]{\textcolor{black!60!black}{#1}}
\setcitestyle{numbers}
\title{A 3D-Printable Dataset for Fair Testing and Comparisons of Tactile Sensors}

\author{
Dexter R. Shepherd$^{1,*}$,
Nicolas Herzig$^{2}$,
Phil Husbands$^{1}$,
Andrew Philippides$^{1}$, \\
Chris Johnson$^{1}$,
William Kimbell$^{1}$,
Oluwaseyi Oladipupo Jesusanmi~$^{1,3}$\\[1ex]
$^{1}$University of Sussex, Department of Informatics\\
$^{2}$University of Sussex, Department of Engineering\\
$^{3}$Sussex Neuroscience, School of life sciences, University of Sussex\\
}
\date{}

\begin{document}

\maketitle

% Abstract (Do not insert blank lines, i.e. \\) 
\abstract{Existing texture datasets for tactile sensing primarily consist of sensor readings from a specific sensor interacting with available surfaces/objects rather than describing the textures themselves, limiting fair comparison between tactile sensors and hindering reproducible research. \corr{In this work, we introduce a six-piece digital dataset of mathematically defined textures designed to be fabricated reliably across different research groups}. The dataset consists of surface patterns derived from combinations of sine-wave and Fourier-based functions, giving controlled variation in spatial frequency, amplitude, and directional structure. \corr{To evaluate the physical repeatability of this framework, we present a validation methodology applied to a six-texture subset.} We evaluate the reproducibility of these textures across three popular 3D printers and filament types in multiple ways. We first use a RGBD D405 stereoscopic camera for scanning the 3D objects to assess difference of the physical print structures. To see how physical differences affect a typical tactile sensor, we next calculate the difference in images captured using an optical TacTip sensor under controlled contact conditions. Our results show that print quality, particularly peak sharpness and stringing, affects tactile variance, with higher-end printers producing more consistent prints. To assess potential downstream effects of the variance, we train two types of classifier, \corr{random forest classifiers} and neural networks, with \corr{PCA preprocessing}, to classify textures. In particular, by training on one batch of textures, but testing on different unseen batches, we further assessed how similar different batches of prints are. These experiments demonstrate that high-quality prints support both strong within-printer generalisation, and cross-printer generalisation. \corr{This work establishes the first openly available, six-texture digital library paired with a proven metrology pipeline, providing an open, modality-agnostic foundation that enables researchers to physically replicate identical textures and fairly compare diverse tactile sensing technologies.}
}

% Keywords  and future work on sim-to-real transfer using matched digital and physical assets
% \keyword{Tactile sensors, 3D printing, Dataset, Haptics, Benchmark, Robotics} 

\section{Introduction}
%Texture classification for comparing and benchmarking tactile sensors remains difficult to reproduce reliably of the textures themselves. 

While texture classification is an important topic in robotics~\cite{lepora2021soft,ward2018tactip,winstone2012tactip,james2018slip,liang2025structure, lima2020dynamic,pratap2024tactile} and used as a benchmark task~\cite{8250220}, fairly comparing texture classification performance between different tactile sensors is difficult due to an inability to reliably reproduce the textures themselves. Existing approaches typically rely on self-collected datasets from available objects~\cite{c2,shepherd2025texture}, which limits the accuracy and fairness of comparisons with newly developed sensors. Indeed, most available texture datasets consist of sensor readings from specific tactile sensors~\cite{lima2023multimodal,marzani2025dataset,zhong2024tactgen}, rather than the textures themselves. This limits their use to model optimisation or transfer learning~\cite{zhao2024transferable}, rather than allowing a direct comparison with other sensors. This paper fills this gap by designing a set of textures that can be 3D printed by researchers, and ensures its physical reproducibility by assessing the variability introduced by different printers and filament types. 

Standardised datasets have been used in other fields. For example 3D printed textures have been employed by Tymms et al. to measure human tactile perception~\cite{tymms2018tactile}. However, this task is aimed at investigating the resolution of human sensing and \corr{consists of similar geometric patterns}. Although it could make a good benchmark for determining the minimal resolution of tactile sensors, it does not evaluate different textures so we require something different for texture classification. In object classification, 3D datasets of objects exist~\cite{zhang2023visual}, which allow the objects to be imported into simulations~\cite{gao2021objectfolder,o2024open}. However, little work has been done with physical instantiations aside from random objects~\cite{cepria2021dataset,8250220}, and Braille~\cite{gao2024enhanced,xu2025neuromorphic,ward2020neurotac}. The latter is a good example of a standardised and reproducible dataset, which is the goal of this paper. 

Following on from the ideas presented in~\cite{tymms2018tactile,gao2024enhanced,xu2025neuromorphic} we propose a 3D printed dataset of textures that can be used to compare the performance of various tactile sensors. To do this we use equation generated texture blocks that can be 3D printed. We then use a TacTip to gather a visual dataset across different printer-filament combinations to determine print variance. Importantly, 3D printing quality varies between different printers and filaments, as well as being affected by other factors that are not always easy to control, such as humidity~\cite{lendvai2024influence}. We therefore investigate how reproducible a dataset is across various popular printers and a range of filament types. \corr{We investigate this variation using three methods}: firstly by scanning repeats of the printed textures using a depth camera and analysing differences; secondly by comparing the differences in the TacTip images from repeat prints; \corr{thirdly}, we train texture classifiers on one repeat and but assess accuracy on the unseen repeats, to assess the impact of dataset variation on classification. In so doing, this work paves the way for fair and reproducible tactile sensor comparisons. %Furthermore, having digital copies of the real world textures could be used for sim-to-real transfer using matched digital and physical assets. 

\section{Methods}

\subsection{Creation of Textures}
We considered various approaches for generating textures in this dataset. Initially, we explored 3D scanning real-world surfaces to better reflect natural textures. However, this introduced variability, including non-uniformity within a single sample. As a result, different regions of the same texture could produce inconsistent sensor responses depending on the contact location, reducing reproducibility and making controlled comparisons difficult.

To address this, we instead draw inspiration from the USAF calibration test~\cite{milstd150index}, which employs well-defined geometric patterns that can be systematically scaled without loss of fidelity. By using mathematically generated patterns, we ensure that each texture is both uniform and precisely controllable. This allows for consistent sampling across the surface, improves reproducibility, and enables more rigorous benchmarking of sensor performance across well-defined stimulus conditions.%Often tactile datasets are uniformed textures~\cite{shepherd2025texture}. 

\corr{The six textures, kept small for practicality, were deliberately designed to share the same fundamental spatial period while differing in their higher-order spectral content and surface topology (e.g., ridged, rounded, square-profile, and point-like asperity structures). This design creates textures that are similar enough to present a meaningful discrimination challenge, while still spanning representative classes of surface geometries encountered in tactile sensing. 
We also intentionally kept the number of textures small to provide a simple, scalable benchmark that is practical to reproduce and extend. This philosophy mirrors established resolution standards in other sensing domains, such as ISO 12233 and the 1951 USAF resolution target~\cite{milstd150index} favouring simplicity, reproducibility, and diagnostic value rather than exhaustive coverage of all possible stimuli.
}
% For this project, we produce six textures, shown in Figure \ref{fig:new-textures-bench}. Each of the 2D surfaces are made from a summation of sin waves. As other repetitive shapes can be obtained from a summation of sin functions, finding the smallest variation in sine patterns that the sensor can detect may help benchmark the sensor for more advanced shapes. %Also, as can be seen from the figure and governing equations, there is a lot of overlap between these equations, making the generation of new patterns that are not widely different easy which is good if one wants to increase the difficulty of classification.
% Each 2D surface is governed by 5 parameters which specify amplitude and frequency along each of the 2 axes as well as a phase shift between the directions. \corr{The selected textures are intended to represent a subset of rigid tactile textures by varying surface geometry (e.g., grooves, bumps, and periodic patterns) and spatial characteristics such as feature size, spacing, and orientation; material-dependent properties such as compliance and friction are outside the scope of this study and are left for future work.} Note that units are not specified at this point which allows the datasets to be printed at different sizes. However, we have only validated one size (specified in section \ref{sec:STL}).

\corr{Following on from the ideas presented in~\cite{tymms2018tactile,gao2024enhanced,xu2025neuromorphic}, we propose a two-phase benchmarking framework to enable the standardised comparison of various tactile sensors. First, we introduce an open-source digital library of six equation-generated texture blocks spanning a highly diversified, discriminative range of structural profiles. Second, we establish a validation methodology to quantify and minimise physical fabrication variance, evaluating six textures across popular 3D printers and filament types. Because 3D printing quality is inherently susceptible to machine-specific errors and environmental variables like humidity~\cite{lendvai2024influence}, we investigate physical dataset reproducibility using complementary methods: 3D optical scanning via a depth camera to map absolute geometric fidelity, comparing the differences of tactile images gathered with a TacTip sensor from the textures and training machine learning classifiers on the visual tactile datasets gathered.  In doing so, this work provides both a highly diversified digital benchmark asset and an empirical roadmap for achieving high-fidelity, cross-lab physical replication.}
%Indeed, as other repetitive shapes (squares, pyramids, etc..)  can be obtained from a sum of sine functions, detecting what is the minimum sine pattern that the sensor can detect may help benchmark the sensor for more advanced shapes. 

\corr{The equations used to generate the textures are given below, representing a wide range of scalable surfaces. Some textures are deliberately similar to make discrimination between them challenging, a desirable property of a test suite for benchmarking sensors and classification techniques.}   

\begin{figure}[H]
    \centering
    \subfloat[Texture 1]{%
  \includegraphics[clip,angle=0,width=0.25\columnwidth]{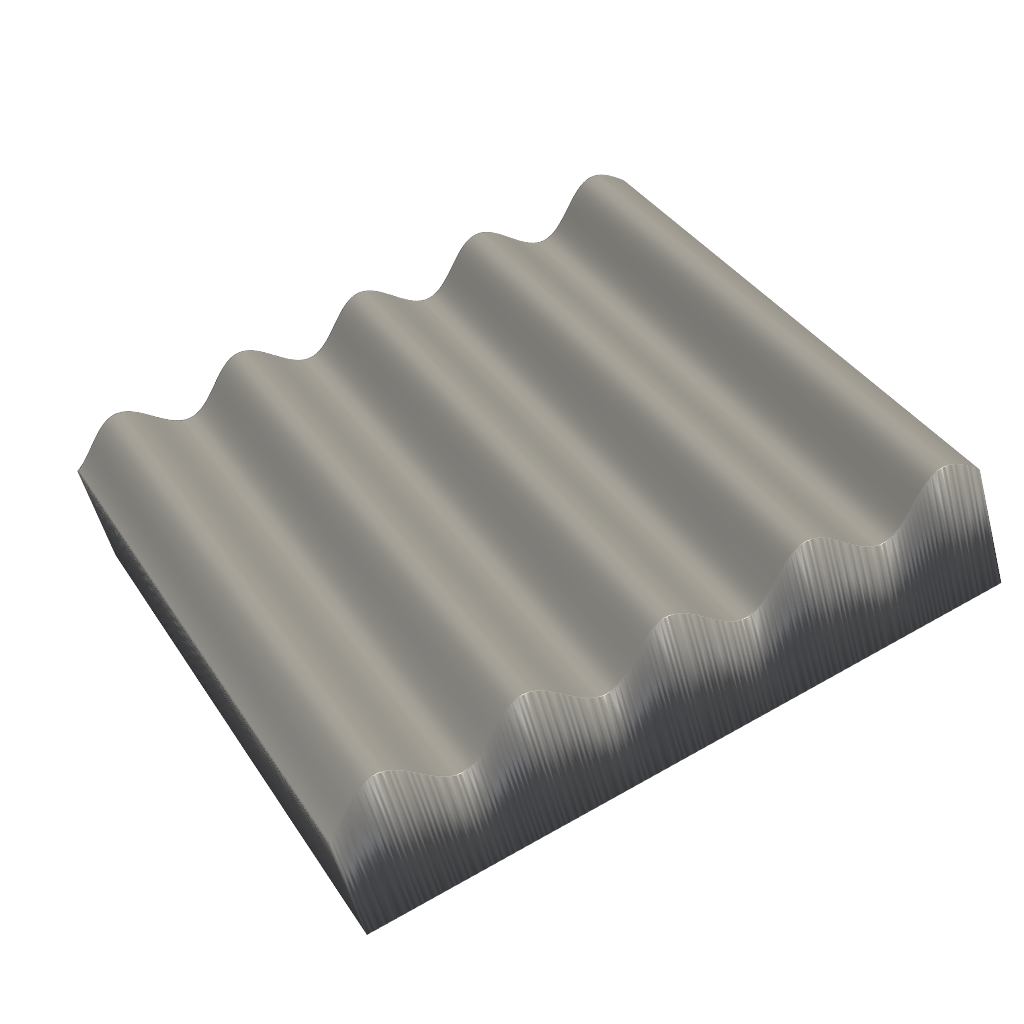}%
}
\subfloat[Texture 2]{%
  \includegraphics[clip,angle=0,width=0.25\columnwidth]{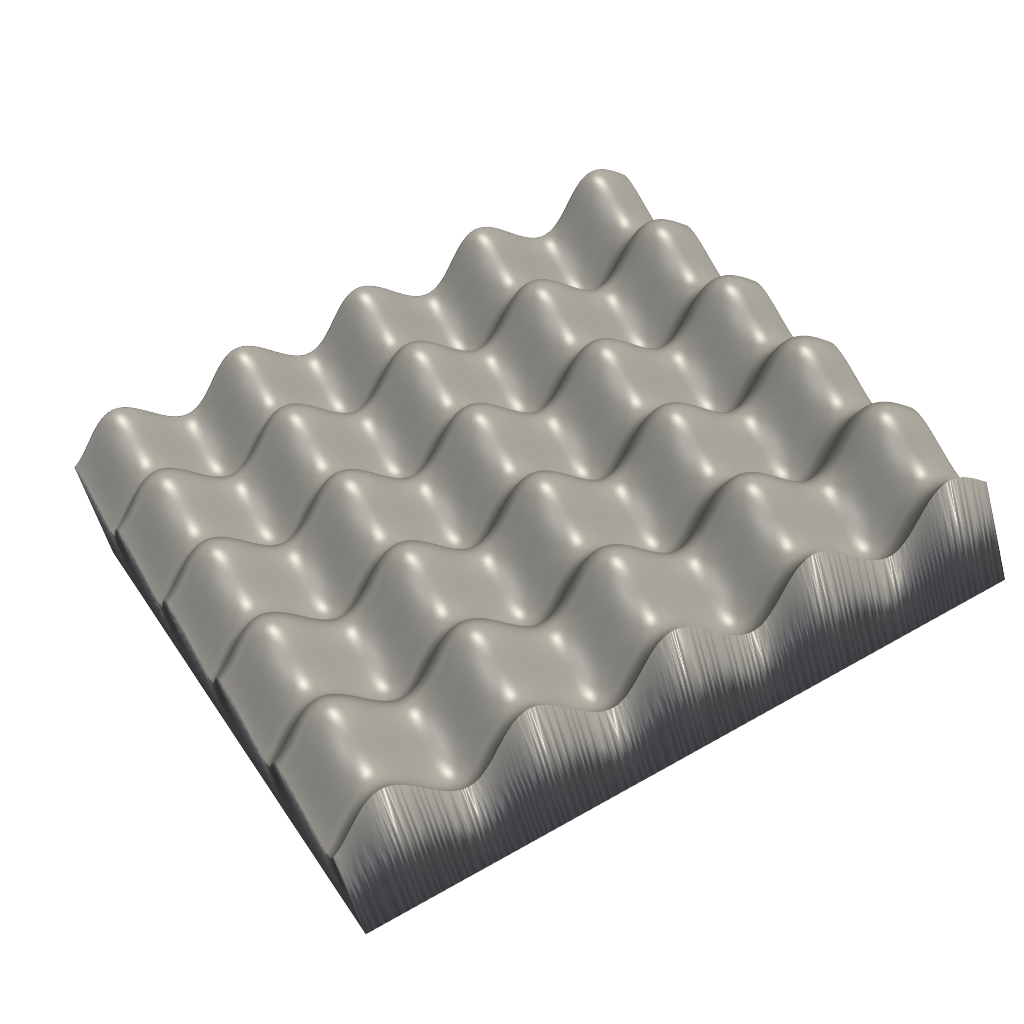}%
}
\subfloat[Texture 3]{%
  \includegraphics[clip,angle=0,width=0.25\columnwidth]{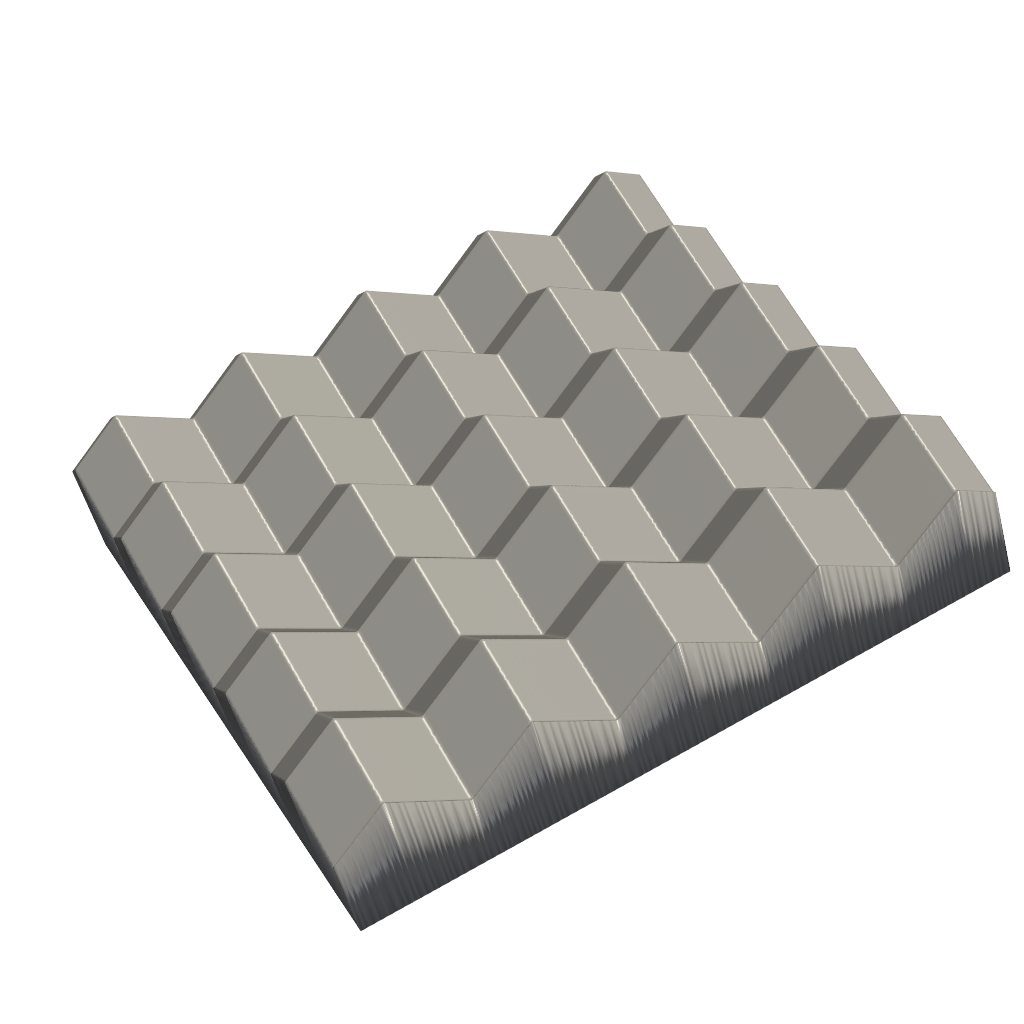}%
}
\\
\subfloat[Texture 4]{%
  \includegraphics[clip,angle=0,width=0.25\columnwidth]{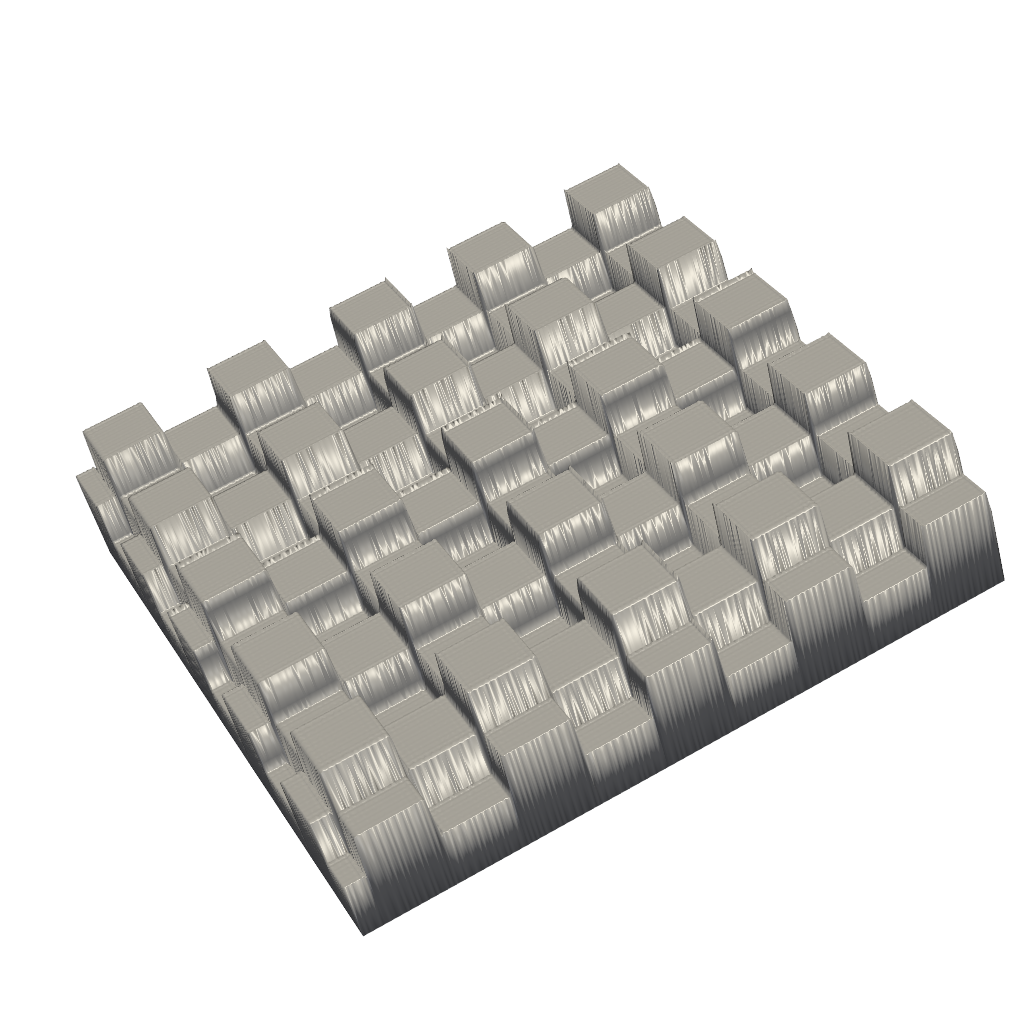}%
}
\subfloat[Texture 5]{%
  \includegraphics[clip,angle=0,width=0.25\columnwidth]{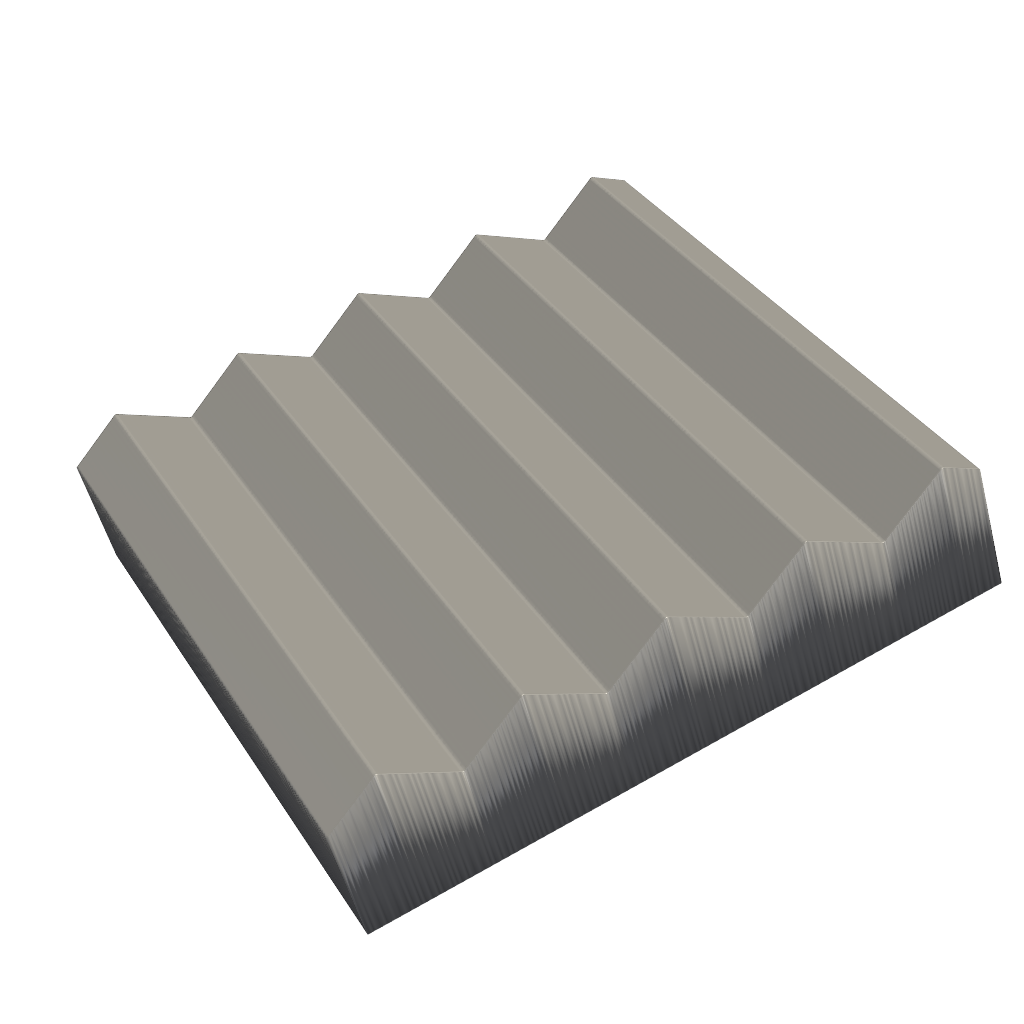}%
}
\subfloat[Texture 6]{%
  \includegraphics[clip,angle=0,width=0.25\columnwidth]{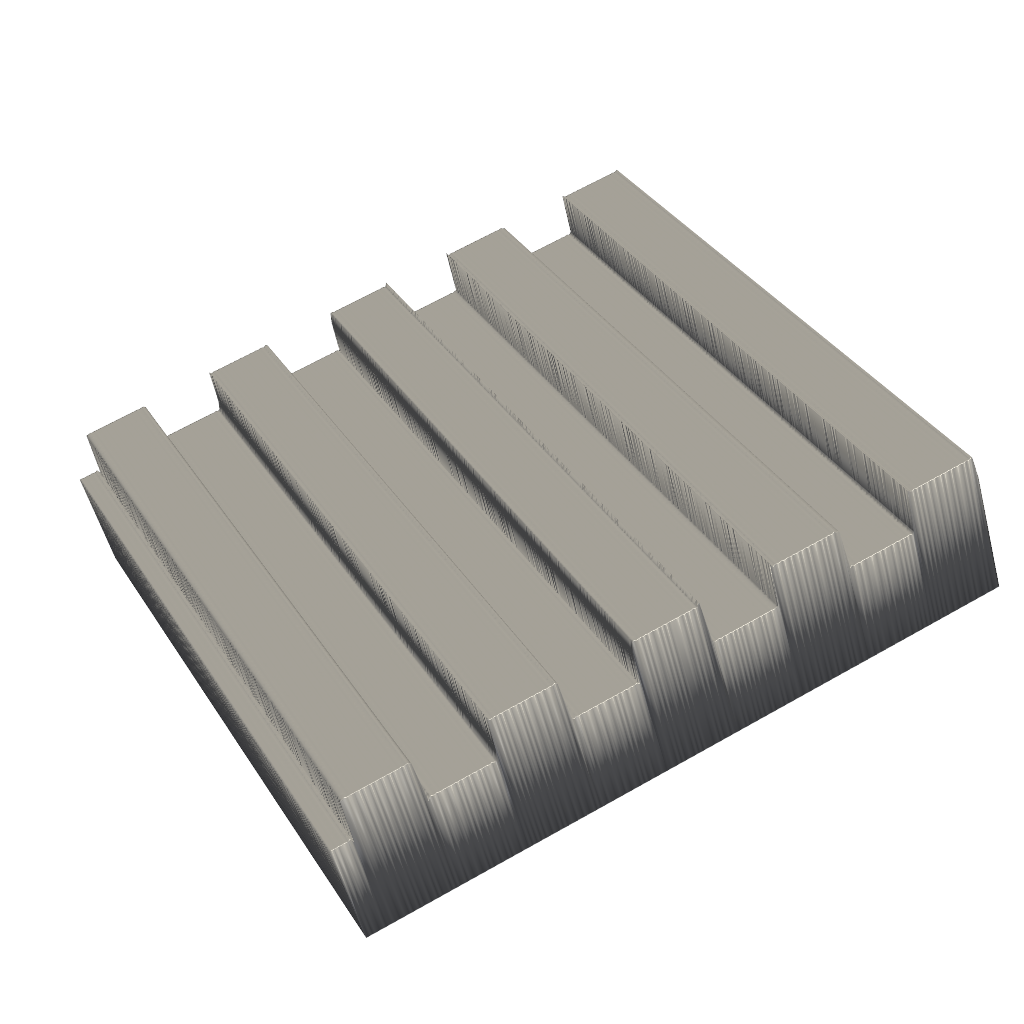}%
}
    \caption{\corr{The 3D printable textures generated from the equations listed above. %We can see that there is a lot of overlap between these textures equations, making the generation of new patterns that are not widely different very possible.
    } }
    \label{fig:new-textures-bench}
\end{figure}

\paragraph{\textbf{Textures One and Two}}
Textures One and Two are generated using equation~\ref{eq_1}. The parameters \( A_1 \) and \( A_2 \) represent the amplitudes of sine waves in the \( x \)- and \( y \)-directions, respectively, \( f_1 \) and \( f_2 \) denote the frequencies of the sine components along the \( x \)- and \( y \)-axes and \( \phi \) is a phase offset applied to the sine term involving \( y \). The variables \( x \) and \( y \) are spatial coordinate inputs, and the output \( z_1 \), the sum of two sine functions modulated by their respective amplitudes, frequencies, and phase, gives the height of the surface. 
\begin{equation}
z_1 = A_1 \sin(f_1 x) + A_2 \sin(f_2 y + \phi)
\label{eq_1}
\end{equation}
For Texture One the parameter values are: $A_1 = 1$, $A_2 = 0$, $f_1 = 1$, $f_2 = 1$, $\phi = \pi / 2$ with $x \in  [0,6\pi+0.5]$ and $y \in [0,6\pi+1.2]$, producing sinusoidal variations in one direction only. The resulting surface is shown in Figure~\ref{fig:new-textures-bench}.

For Texture Two the parameter values are: $A_1 = 1$, $A_2 = 1$, $f_1 = 1$, $f_2 = 1$, $\phi = \pi / 2$ with $x \in  [0,6\pi+0.5]$ and $y \in [0,6\pi+1.2]$, introducing sinusoidal variation in both dimensions. %Such feature introduces variance for the potential of robustness checks. 

\paragraph{\textbf{Texture Three}}
For Texture Three we use the function \( z_3 \) defined over a grid using the following summation (Equation~\ref{eq_2}):

\begin{equation}
z_3 = \sum_{\substack{i=1 \\ i \text{ odd}}}^{N} \left(A_i \cdot \sin(i x) + A_i \cdot \sin(i y) \right)
\label{eq_2}
\end{equation}
\noindent In this expression, \( N \) is the number of terms in the summation, taken over odd integers from 1 to \( N \). For our texture we used $N=25$. Each term in the series uses the coefficient \(A_i\) which controls the amplitude of the sine wave and is defined by Equation~\ref{eq_2.5}.

\begin{equation}
A_i = \frac{8}{\pi^2} \cdot \frac{(-1)^{\frac{i - 1}{2}}}{i^2}
\label{eq_2.5}
\end{equation}
\noindent  The sine functions \( \sin(i x) \) and \( \sin(i y) \) vary with spatial coordinates \( x \) and \( y \), respectively, and share the same frequency \( f = i \) and amplitude \( A_i \). The ranges of \( x \) and \( y \) are the same as for Textures One and Two. These ranges are used for all the textures described in this section.  

%\( z_3 \) is initialized as a zero-valued array of the same shape as \( x \), and is built-up by adding sine wave components in both the \( x \)- and \( y \)-directions. This results in a 2D Fourier-like series that combines symmetrical sine waves to construct a spatial pattern.

As \( z_3 \) is built-up by adding sine wave components in both the \( x \)- and \( y \)-directions, the result is a 2D Fourier-series-like symmetrical pattern (see Figure~ \ref{fig:new-textures-bench}).

\paragraph{\textbf{Texture Four}}
Texture Four uses the same equation as Texture Three (Equation~\ref{eq_2}), but with a modified amplitude coefficient, \(A_i\), given by Equation ~\ref{eq_4.5}.

\begin{equation}
A_i = \frac{4}{\pi i}
\label{eq_4.5}
\end{equation}

\noindent The decay and lack of sign alternation lead to sharper features, with more pronounced ridges and edges (see Figure~ \ref{fig:new-textures-bench}).

\paragraph{\textbf{Texture Five}}
Texture Five is generated by the function \( z_5 \) shown in Equation ~\ref{eq_5}. 

\begin{equation}
    z_5 = \sum_{\substack{i=1 \\ i \text{ odd}}}^{N} \left(A_i \cdot \sin(i x) \right)
\label{eq_5}
\end{equation}

Where \(A_i\) is defined by Equation ~\ref{eq_2.5} (as for Texture Three). With variation along one axis only, the shape has an extruded profile (see Figure~ \ref{fig:new-textures-bench}).

\paragraph{\textbf{Texture Six}}

Texture six uses the same function as Texture Five (equation~\ref{eq_5}), but with the coefficient, \(A_i\), now defined by  equation~\ref{eq_4.5}, as in Texture Four.

\subsection{Conversion to STL}\label{sec:STL}
All the models were converted to file type STL as blocks to be 3D printed. These models were all constrained to be the same maximum height of 7mm as we wanted to ensure the sensor can be lowered to the same position for each experiment. When generating the patterns shown in Figure~\ref{fig:new-textures-bench}, we used a step size of 0.1 which converts to 0.53mm per step, with 189 points in the x-direction and 201 points in the y-direction so that the blocks are not perfectly square, giving blocks of dimension 23.3mm $\times$ 30mm. This was done to prevent human error and ensure the blocks only fit in the testing rig one way. These points form the vertices \corr{of the triangulated STL mesh}. The smaller the step size, the larger the STL file and the higher \corr{the geometric fidelity of the exported model}. \corr{The step size was selected such that the resulting mesh discretisation error (e.g., chordal deviation) remained below the nominal manufacturing resolution of the printers listed in Table~\ref{table_print_specs}. Consequently, any observed limitations in surface smoothness are attributable to the manufacturing process rather than the STL export. For FDM printing, increasing the STL resolution beyond this point provides no practical benefit as the achievable geometry is ultimately constrained by the nozzle diameter and layer height}. The same resolution constraints do not apply to the resin printer. However, from prototype prints, a step size of 0.1 showed to be smooth upon visual inspection, so we did not need to increase resolution. 

\begin{figure}[H]
    \centering
    \includegraphics[width=0.5\linewidth]{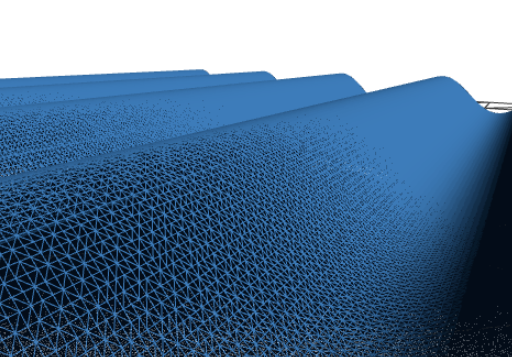}
    \caption{Example wireframe of the STL file for texture 1 generated by equation~\ref{eq_1}. We used step size 0.1, which gave more than high enough resolution given the constraints of nozzle size of the printers.}
    \label{fig:wireframe}
\end{figure}

\subsection{Tactile Dataset Printing and Collection}
\subsubsection{Experimental Setup}
\begin{figure}[H]
    \centering
    \includegraphics[width=0.5\linewidth]{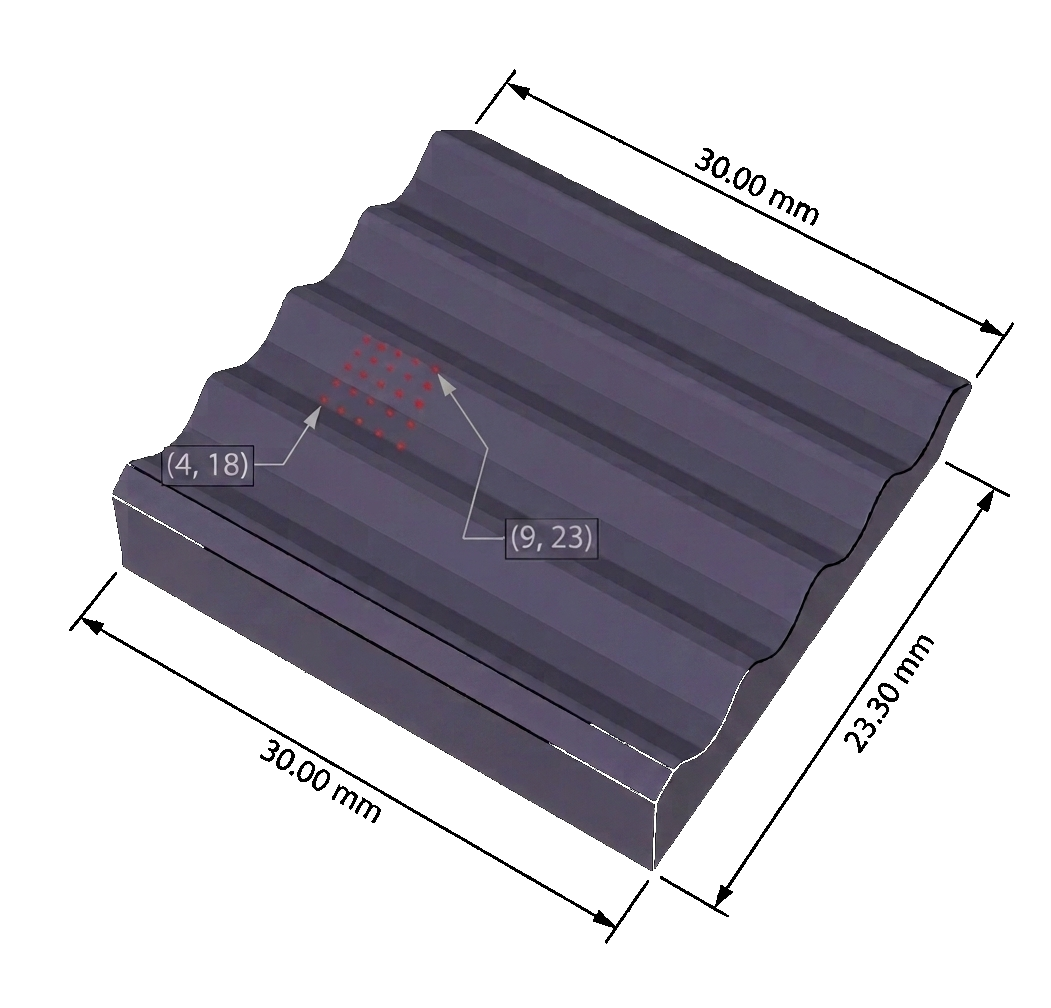}
    \caption{A diagram to show an example block, and the positions that the readings have been taken from in the 5 $\times$ 5 grid. Each red point is where the lowest point of the TacTip soft body will press. }
    \label{fig:griddes}
\end{figure}

\begin{figure}[H]
    \centering
    \subfloat[Testing rig]{%
     \includegraphics[width=0.5\linewidth]{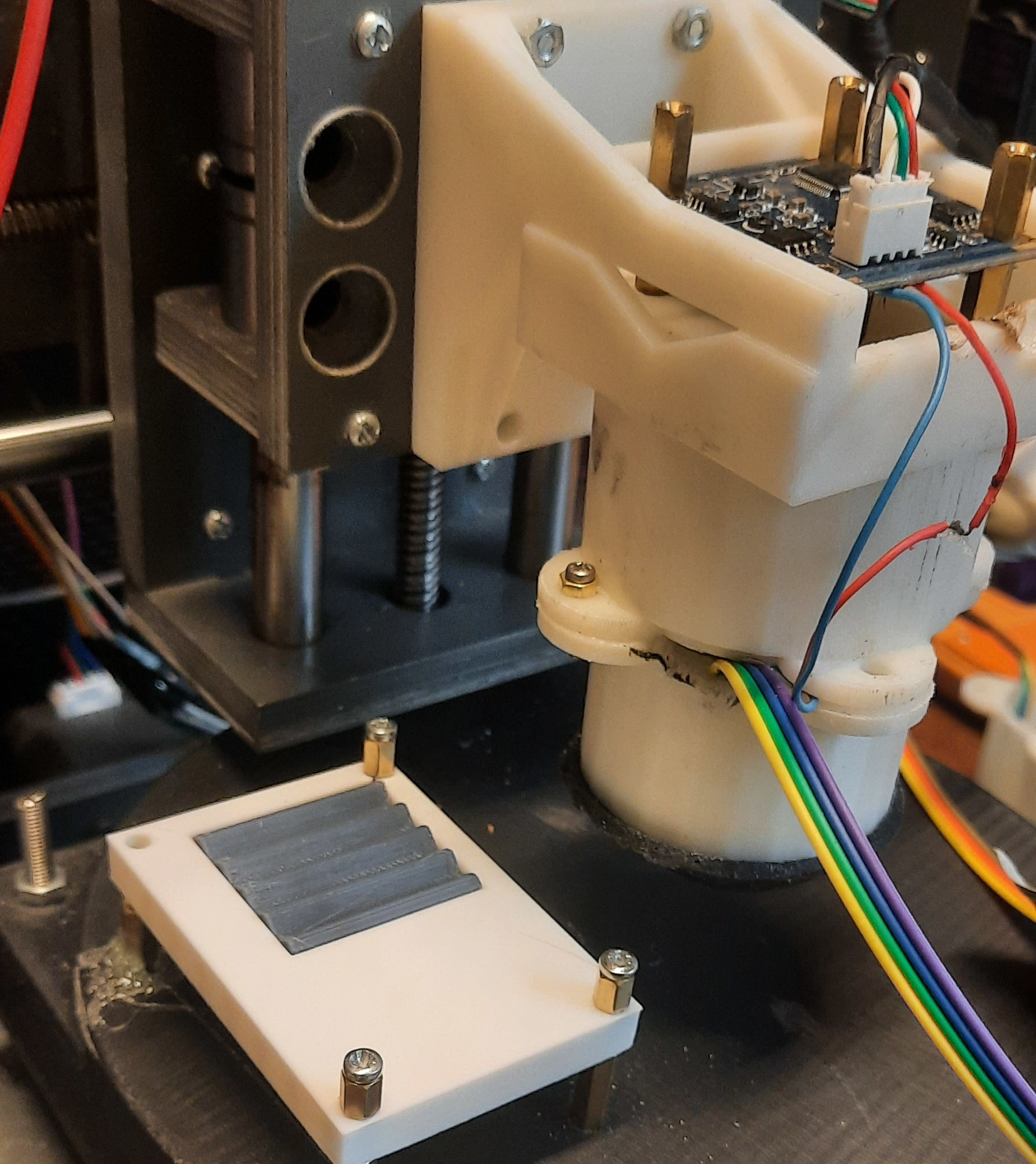}
    }
   \subfloat[Printing setup]{%
     \includegraphics[width=0.5\linewidth]{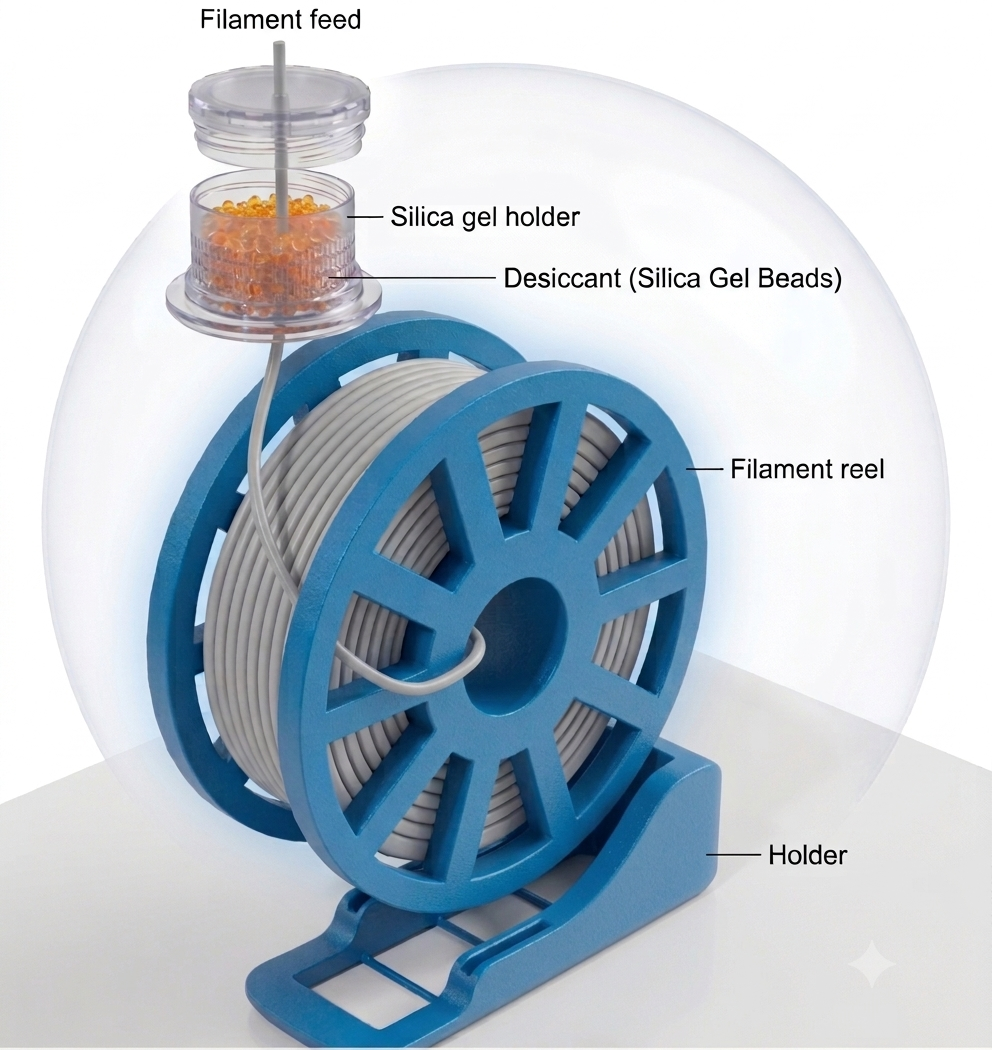}
    }
    \caption{\emph{a}: Rig with new texture housing unit attached. Each texture slots into the unit, allowing for the sensor to be pressed. \corr{\emph{b}: Render of the reel housing unit for storing the filament in a non-humid environment. For visualisation the outer housing is not shown, but the reel is inside an air tight container.}  }
    \label{fig:texturehousing}
\end{figure}

To gather the dataset, we used a TacTip tactile sensor~\cite{lepora2021soft} with 133 1mm optical markers pressed onto a grid of positions across the texture. Figure~\ref{fig:griddes} shows the points of contact for the TacTip sensor. Each sensor is pressed twice on the contact points. A texture housing unit (seen in Figure~\ref{fig:texturehousing}a) was 3D printed to hold each of the texture blocks seen in Figure~\ref{fig:new-textures-bench}. The testing rig pressed the sensor over each texture by pressing the sensor in a rectangular grid taking snapshots of the tactile sensation. The x position of the centre of the TacTip was moved to 5 equally spaced positions between 4mm and 9mm, and the corresponding y position was moved to 5 equally-spaced 1mm positions between 18mm and 23mm, giving 25 points across the texture. This was repeated twice giving a dataset of 50 images. Previous experiments have used varying force levels~\cite{shepherd2025texture,eguchi2026cluster,wang2022fabric}), however as shown in the non-linear classification studies of previous work~\cite{shepherd2025texture}, the classification is invariant to force. Therefore we only used one sensor contact force for our dataset at 1.9 $\pm$ 0.009 N. 
%training algorithm

\subsubsection{3D printers}
We selected four printers, chosen for popularity and access: An Ender-3 printer; a Creality Ender V3-SE printer; a BambuLab P1P printer; a Formlabs Form 3 resin printer. \corr{Each printer type was tested with a single representative printer of that model, as we did not have access to more than one of each type}. The printer specifications are outlined in Table~\ref{table_print_specs}. Resin printers are generally more expensive but yield better quality for small detail. Some labs will have different printers to these, so replicating our methods will allow researchers to see whether their printer can produce a high enough standard. We used Resin Grey v4.1 to create the textures for this study. 

\begin{table}[H]
    \centering
    \begin{tabular}{|l|p{2.3cm}|p{2.3cm}|p{2.3cm}|p{2cm}|}
    \hline
    \textbf{Spec} & \textbf{Ender-3} & \textbf{Creality Ender-3 V3 SE} & \textbf{Bambu P1P} & \textbf{Formlabs Form 3 (Resin)}\\
    \hline
    Build volume & 220$\times$220$\times$250 mm & 220$\times$220$\times$250 mm & 256$\times$256$\times$256 mm  & 145$\times$145$\times$ 185 mm\\
    \hline
    Motion system & Cartesian Bed slingers & Cartesian Bed slingers & CoreXY  & na \\
    \hline
    Acceleration & $\leq$20,000 mm/s$^2$ & $\sim$2,500 mm/s$^2$ & 20,000 mm/s$^2$ & na\\
    \hline
    Hotend temp & up to 300$^\circ$C & up to 260$^\circ$C & up to 300$^\circ$C & na\\
    \hline
    Bed temp & $\leq$100$^\circ$C & $\leq$100$^\circ$C & $\leq$100$^\circ$C & na \\
    \hline
    Extruder & Direct drive & Sprite direct drive & Direct drive, all-metal & na\\
    \hline
    Extruder aperture & 0.4 mm & 0.4 mm & 0.4 mm & na\\
    \hline
    Leveling & Auto (varies by model) & CR Touch + strain & Auto (built-in) & na \\
    \hline
    Connectivity & Touchscreen, SD/USB & Touchscreen, SD/USB-C & App, cloud, slicer & USB\\
    \hline
\end{tabular}
\caption{Comparison of Ender-3, Creality  Ender-3 V3 SE, Bambu Lab P1P printers and Formlabs Form 3 resin printer. It is worth noting that the printers all used 0.4 nozzles, though this can be reduced further. }

\label{table_print_specs}
\end{table}

We used three different 1.75 mm filaments: Rapid PETG (Elegoo), PLA+ (eSUN), and PLA-Lite (eSUN). For the resin printer we used Grey V4. According to manufacturer specifications, these filaments have a dimensional tolerance of $\pm$ 0.02 mm. We chose the PLA filaments as PLA is an economic standard across many 3D-printer users. The PLA-Lite (we refer to as PLA-) was cheaper than the PLA+, though the PLA+ was supposed to be higher quality so we chose to explore whether this had an impact. We used PETG as it provides durability but can be trickier to print with, potentially introducing more variance unless print settings are tuned properly.

\subsubsection{Humidity and Environment Control}
\corr{To minimise the effects of moisture absorption during filament storage and to ensure consistent print quality, all filament spools were stored in sealed dry containers based on the \emph{Ultimate Filament Storage} cereal box design from BambuLabs~\cite{ultimatefilamentstorage}. Each container incorporated dedicated desiccant compartments and an integrated hygrometer, allowing the internal humidity to be monitored while maintaining a low-moisture environment. The sealed containers reduced the exposure of hygroscopic filaments to ambient humidity between printing sessions, helping to prevent moisture-related defects such as bubbling, stringing, and reduced inter-layer adhesion. This storage approach ensured that variations in print quality caused by filament moisture content were minimised, improving the repeatability and consistency of the printed components used throughout this work. The render of this container can be seen in Figure~\ref{fig:texturehousing}b. Prior to the filaments being stored in the container, they were put in a filament dryer, PLA for 8 hours at 50$^{\circ}C$ and PETG for 60$^{\circ}C$ for 8 hours. Both were then immediately placed in the containers. }

\corr{In terms of temperature and humidity control during the printing process, the Form 3 resin printer operates within an enclosed and temperature-regulated environment, reducing the influence of ambient conditions on print quality. All printers were located within the same laboratory space, where humidity was managed using an air purification system. Consequently, while environmental factors cannot be entirely eliminated, the printing conditions were kept consistent across all experiments.} 

% No additional filament drying equipment was used with the Ender-3 printer, as the intention was to evaluate a representative low-cost consumer FDM setup that reflects the configuration commonly available to researchers and hobbyists.

\subsubsection{Scalability and Hardware Limitations}
\corr{While the benchmark geometries are saved as unitless STL files, making the underlying mathematical mesh inherently scalable to any dimension, the physical instantiation of these textures remains bounded by the minimum feature size and resolution of the specific 3D printer used. This study validates the benchmark at a single physical scale. Scaling the textures down significantly may result in the loss of high-frequency geometric features (such as wave peaks) if they fall below the printer's manufacturing limits, while scaling up may alter the mechanical interaction dynamics with the tactile sensor. }

\subsection{Evaluation of Materials and Print Quality}
 We wanted to evaluate the reproducibility of textures produced by different 3D printers and whether different filament brands and types differ. If a cheaper filament, or printer, is significantly worse in quality than others, then we need to advise against using this, or find combinations of print settings mitigating these issues. 
\corr{We employed two methods of evaluation, one using an RGBD Intel Realsense D405 stereoscopic camera to assess structural differences, and the other using a Tactile sensor (TacTip) to investigate the variance in sensor response and how this variance impacted classification tasks}.

\subsubsection{3D Scan Evaluation}
\label{scanning}
\begin{figure}[H]
    \centering
    \includegraphics[width=1\linewidth]{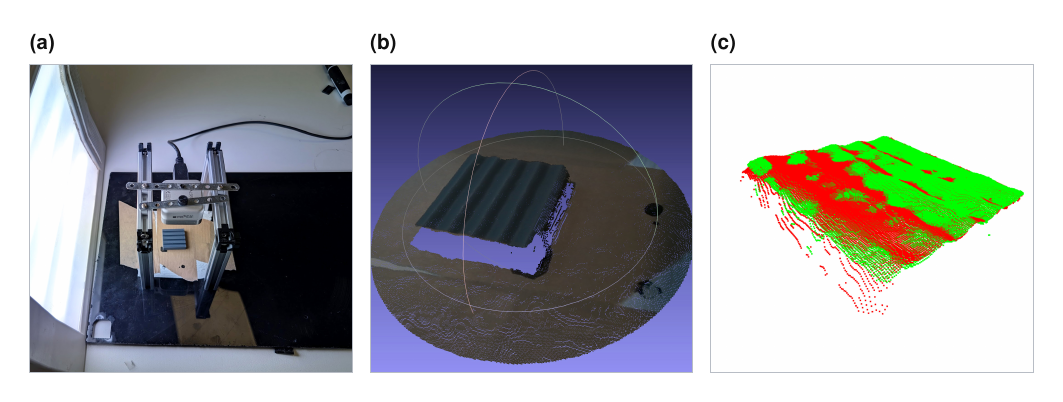}
    \caption{(a): Scanning rig for capturing 3D point clouds of the texture blocks. Top lamp not shown here for clarity. (b) Visualisation of captured point cloud of a texture block after cropping. (c) Alignment process during the comparison of two texture block. Red is one texture block scan and green is the other.}
    \label{fig:scan methods}
\end{figure}

\corr{For each, printer, filament and texture combination that we investigated in detail, we made five repeat prints. We then used the RGBD Intel Realsense D405 stereoscopic camera to scan the texture blocks in 3D. The camera has a scanning precision of < 1 mm when at its ideal recording range. We used Intel Realsense viewer software to record at a resolution of 848x480 and export the scans to PLY format (further software recording parameters are provided in the Appendix within tables~\ref{tab:d405_config}\&\ref{tab:d405_parameters}). We recorded the texture blocks from a top-down perspective at a range of 8.5 cm, with supplementary lighting provided by lamps on the side and top of the scanning rig (setup shown in Figure~\ref{fig:scan methods}a). 
These scans were static from only one perspective, so aspects of the texture blocks side or underside were not captured. }

% To align the point clouds before comparison we used an iterative closest point (ICP) algorithm. For texture comparison we compared the point clouds of texture blocks printed using the same material and texture type, in order to evaluate extent of any error of the print process. We compared the texture scans by measuring the sum of distances between respective points in clouds. Measuring the difference between multiple scans of the same block provided a metric for scanning error. Measuring the difference between scans of different blocks of the same texture type and materials gave us a metric for print error.}

We saved each scan as a coloured 3D point cloud (Figure~\ref{fig:scan methods}b), then filtered the point cloud so that only points from the texture remained (Figure~\ref{fig:scan methods}c). Specifically, each scan was preprocessed to remove scan noise by finding the largest cluster of points (which would be what the camera captures directly in front of it). The centroid point of this cluster is calculated, and a 50mm radius is cropped (as seen in Figure~\ref{fig:scan methods}b). Finally we remove everything under the z-axis threshold of 2mm from the z-axis centroid to only keep the top of the point clouds, as seen in the two point clouds in Figure~\ref{fig:scan methods}c. 

The processed point clouds were aligned using an iterative closest point~\cite{arun1987least} (ICP) registration algorithm. Floor regions were first removed through RANSAC plane segmentation using a distance threshold of 2 mm. A multi-stage point-to-point ICP procedure was then applied using progressively smaller voxel sizes and correspondence thresholds. This coarse-to-fine registration approach initially corrected larger positional offsets before refining the alignment to achieve a high-resolution fit and results in a transformation matrix $T^{*}$ as follows: 

\begin{equation}
    T^{*} = \arg\min_{T} \sum_{i} \left\| p_i^{\text{target}} - T\left(p_i^{\text{source}}\right) \right\|^2
    \label{pointclouddiff}
\end{equation}
\noindent where $p_i$ is a position of a point in either the target or source point-cloud and $T$ is a candidate transformation. Because this algorithm is based on minimising the distance between the point clouds, the final transformation matrix, $T^{*}$, obtained from ICP is applied to the source point cloud and the mean distance, d, across all points in the point cloud is used as the distance between two repetitions of a printed texture. 

To calculate how much variance is introduced for a particular texture by the printer-filament combinations specified in Table \ref{tab:printer_errors_scan}, we compare the mean distance between each combination of repeats (i.e. repeat 1, compared to repeats 2 to 5; repeat 2 compared to repeats 1 and 3 to 5; etc). As the ICP algorithm is assymetric in how it treats target and source, this gives 30 comparisons in total for each texture and printer-filament combination. We call this the \textit{raw error} and present the mean and standard deviation across the comparisons as well as the number of comparisons in Table \ref{tab:printer_errors_scan}.

As well as the raw error we also report what we call the \textit{method error}. This is calculated using the procedure above, but the comparisons are made from five scans of the \textbf{same} block, across all six textures and each printer-filament combination. This gives 20 comparisons in total for each texture and printer-filament combination and we again present the mean and standard deviation in Table \ref{tab:printer_errors_scan}. The resulting differences show how much the points differ due to LED lighting, print surface reflection and camera noise. These noise variations causing error within the predictions are referred to as the noise floor. By statistically comparing the distribution of distances from repeated prints of a texture, with repeated scans of the same block, we can see whether the signal cuts through the noise floor.  To do this we use a t-test and report the p-value in Table \ref{tab:printer_errors_scan}.

\subsubsection{Tactile Image Evaluation}
\label{section:imageEval}
To assess how much images from the TacTip varied, we calculated the differences between images gathered from a TacTip sensor pressed into the five repeat prints (from each texture, printer-filament combination), as pressing the sensor on the same part of the block should result in similar sensor readings. For each print, we press the TacTip twice on 25 positions giving 50 images per print. We then calculate the mean-squared difference between all pixels:

\begin{equation}
Imdiff(x,y,p,q,a,b) = \frac{\sum^{n-1}_{i=0}(Im(x,y,press_p,repeat_a)_{i}-Im(x,y,press_q,repeat_b)_{i})^2}{n}
    \label{MSE_DIFF}
\end{equation}

\noindent where $Im_i$ is the i'th pixel in an image and we calculate the mean across the $n$ pixels. For the image, x and y are coordinates of the positions, $press_p$ denotes which of the two presses is used (so $p\in{1,2}$) and $repeat_a$ is which repeat of the block is compared with $a\in{1,6}$ for most conditions, though  $a\in{1,5}$ for two (see below). To calculate the mean difference between repeats, we calculated Imdiff using equation \ref{MSE_DIFF} for unique combinations of repeats (i.e. repeat 1 vs repeats 2 to 5; repeat 2 vs repeats 3 to 5 etc) for each of the 25 positions (x,y) and each of the two presses (so $p=q$ in equation \ref{MSE_DIFF}). Where there were five repeats, this gives $25\times 2 \times (5+4+3+2+1)=750$ differences, and 500 differences when there are 5 repeats. 

To capture the noise introduced by fluctuations in LEDs, positioning etc, we also calculated a method error for the tactile image comparison. To do this, we calculated differences between each press for a given position for the same block i.e.: Imdiff(x,y,1,2,a,a) where $a\in{1,6}$. This resulted in either $25\times 6=150$ or $25\times 5=125$ differences. We also gathered a recording of the TacTip stationary, not pressed on anything, to assess the component of error from LED fluctuation. We calculate the MSE between 136 frames of recording to calculate the background noise, as in Equation~\ref{MSE_DIFF}, between each frame. The average fluctuation is 3.67 which sits far below the difference readings shown in Section~\ref{imageres}.
%For the background noise measurement we have $n=136$ calculated from approximately 10 seconds of video. 

 %If there is insignificant difference in variance between datasets then we can assume that the print quality makes no difference to the model. 
%
We tried to get readings from five repeats for each texture and printer-filament combination. However, there were some corruptions in the dataset, where LED power fluctuated and the image was too dark to see anything. Corruption occurred in two samples (Ender 3 using PLA+, referred to as Ender-3 in Tables \ref{tab:printer_errors_scan}-\ref{tab:augmentation_ablation_printers}, Texture 6, repeat 1;  Resin, Texture 1, repeat 1). This meant that these texture and printer combinations had one less set of Tactile images which reduced the overall dataset by 100 images. 

\subsection{Benchmark Reproduction Workflow}
\corr{To encourage reproducibility and facilitate future tactile sensing research, all resources used to construct the benchmark are publicly available through the accompanying GitHub repository. The repository contains the complete pipeline for generating printable texture samples, producing standardised printer-ready files, and reproducing the dataset collection process.}

\corr{The repository is organised into three primary components. First, a library of printable texture models is provided (3DD1), including procedurally generated surface geometries. Secondly, the repository includes scripts for generating custom textures from user-defined parameters (3DD2), allowing researchers to extend the benchmark with additional tactile patterns while maintaining consistent geometric constraints. Finally, standardised printing profiles are supplied to ensure that benchmark samples can be fabricated consistently across compatible material extrusion 3D printers (3DD3). Naming conventions are outlined in Appendix~\label{appendixcode}}. 

\corr{The recommended reproduction workflow consists of four stages. First, users generate or select a texture model from the provided library~\cite{dataset3}. Secondly, the model is sliced using the supplied printer profiles, ensuring identical print settings such as layer height, nozzle diameter, infill, and print orientation. Thirdly, the printed sample is fabricated using the specified filament materials and environmental conditions. Finally, the completed sample can be incorporated directly into the tactile data collection pipeline described in this work.}

\corr{By releasing both the source models and the complete fabrication pipeline, researchers are able not only to reproduce the benchmark presented here but also to generate additional textures that remain compatible with the benchmark methodology. This enables direct comparison between future tactile sensing approaches while promoting consistency across independently collected datasets.}

\section{Results}

\begin{figure}[H]
    \centering
    \subfloat[Texture 1]{%
  \includegraphics[clip,angle=0,width=0.33\columnwidth]{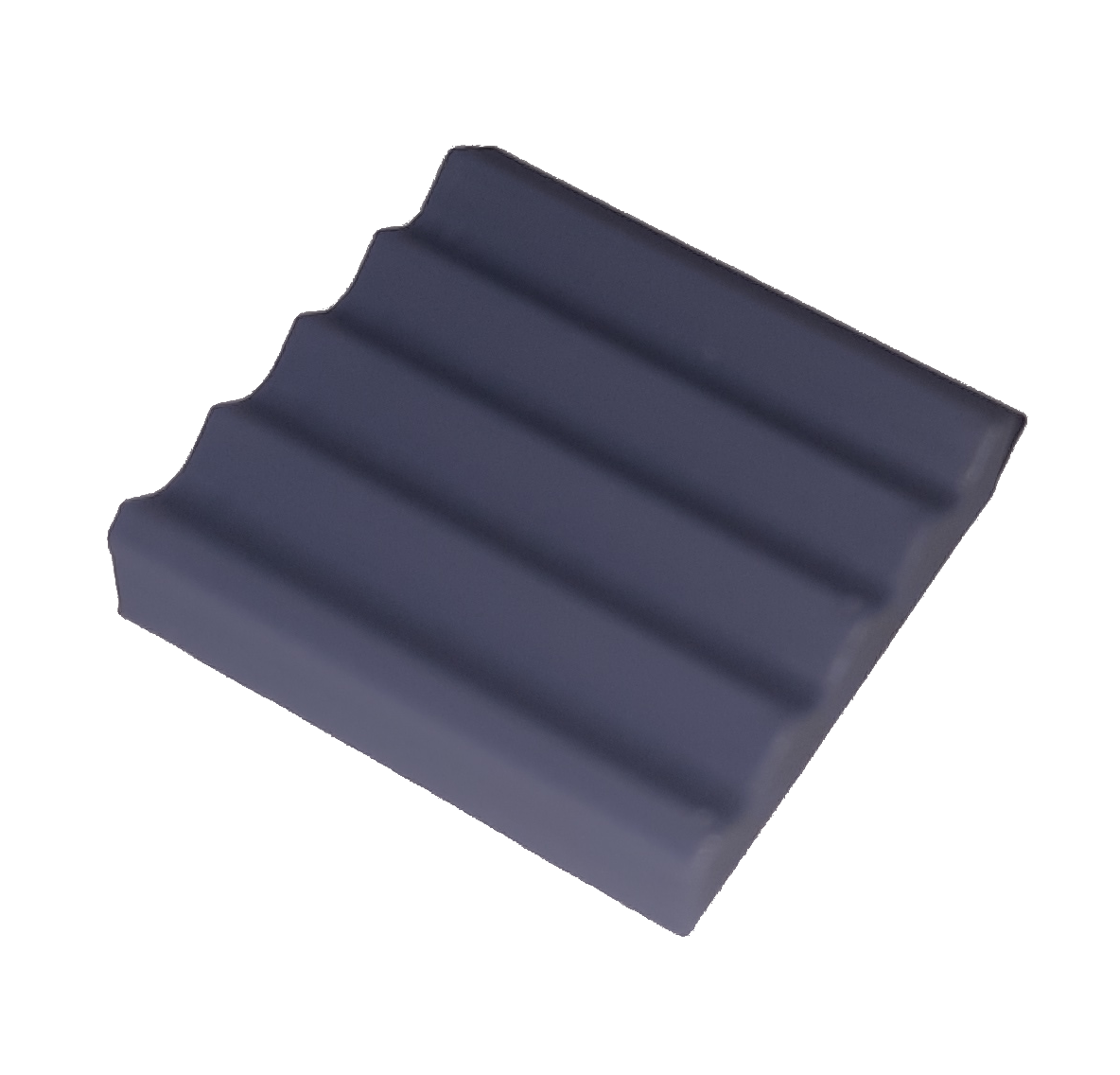}%
}
\subfloat[Texture 2]{%
  \includegraphics[clip,angle=0,width=0.33\columnwidth]{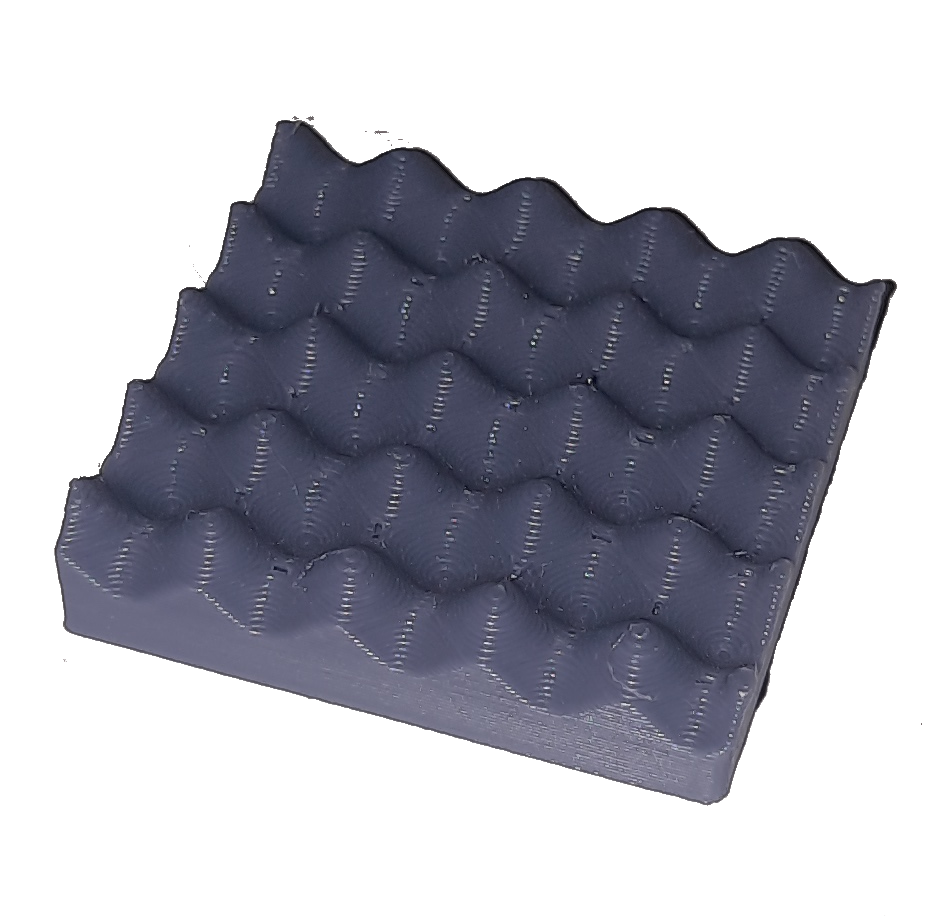}%
}
\subfloat[Texture 3]{%
  \includegraphics[clip,angle=0,width=0.33\columnwidth]{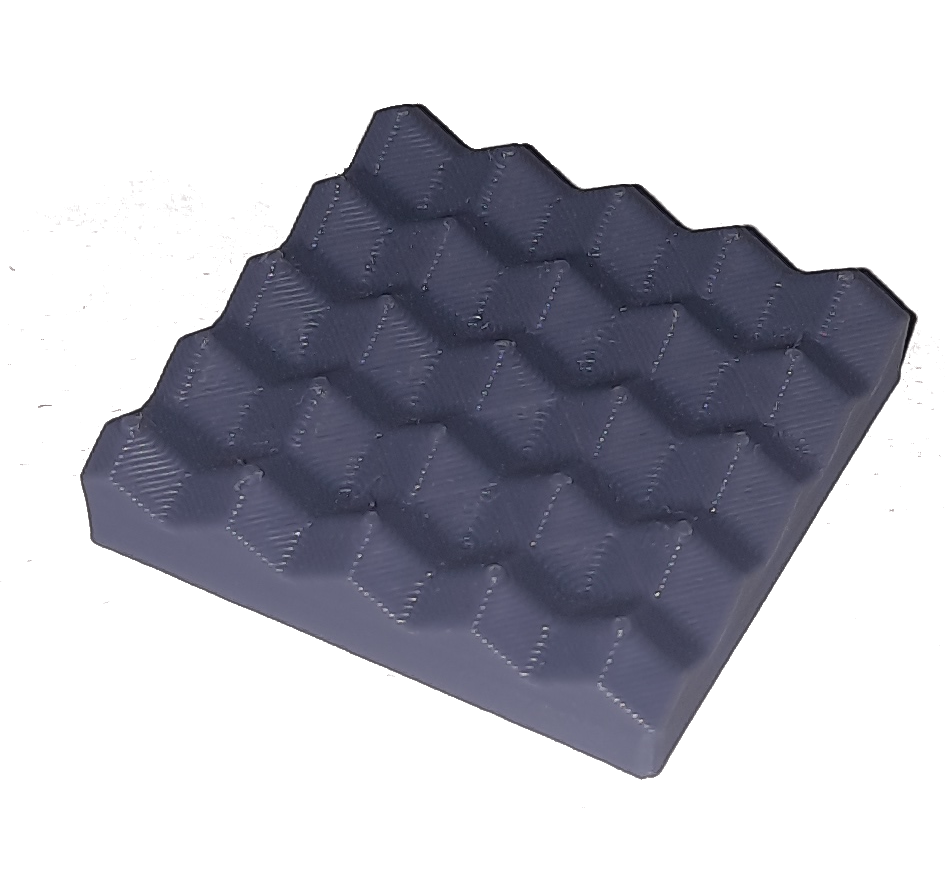}%
}

\subfloat[Texture 4]{%
  \includegraphics[clip,angle=0,width=0.33\columnwidth]{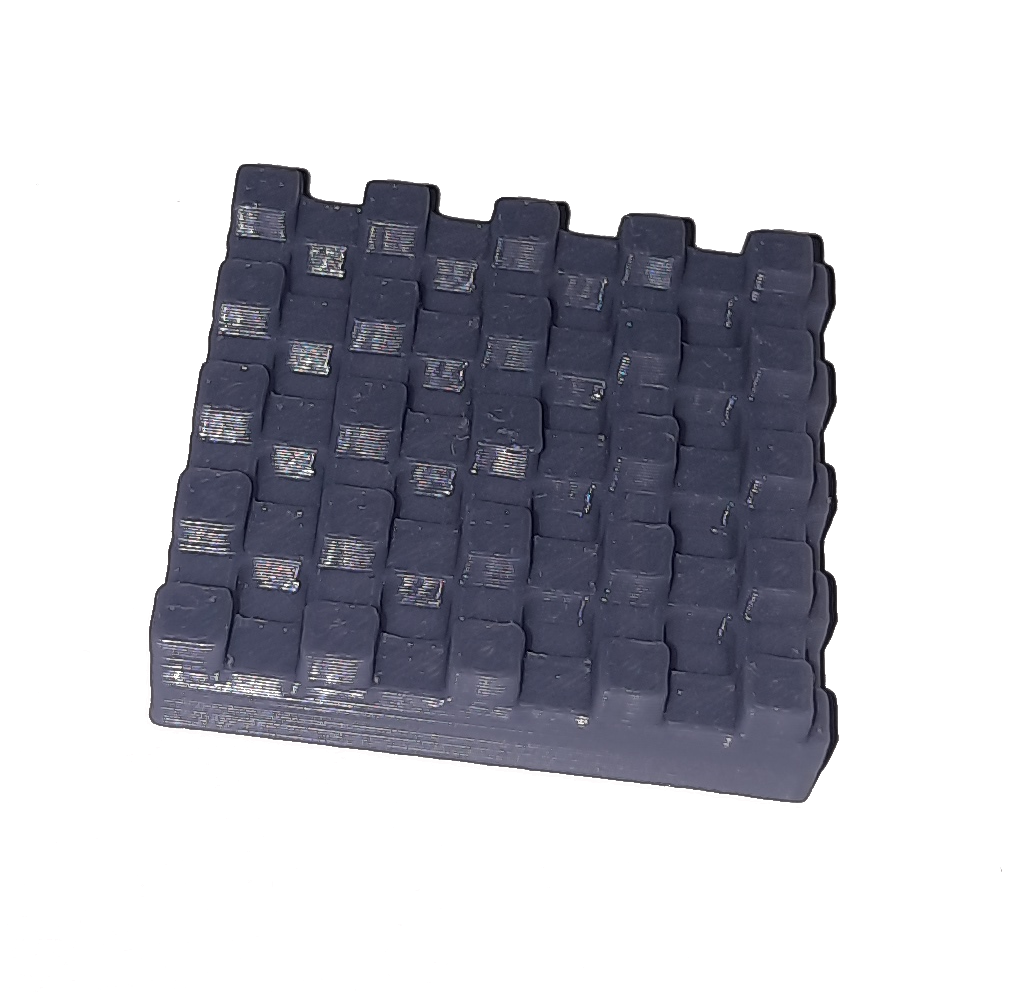}%
}
\subfloat[Texture 5]{%
  \includegraphics[clip,angle=0,width=0.33\columnwidth]{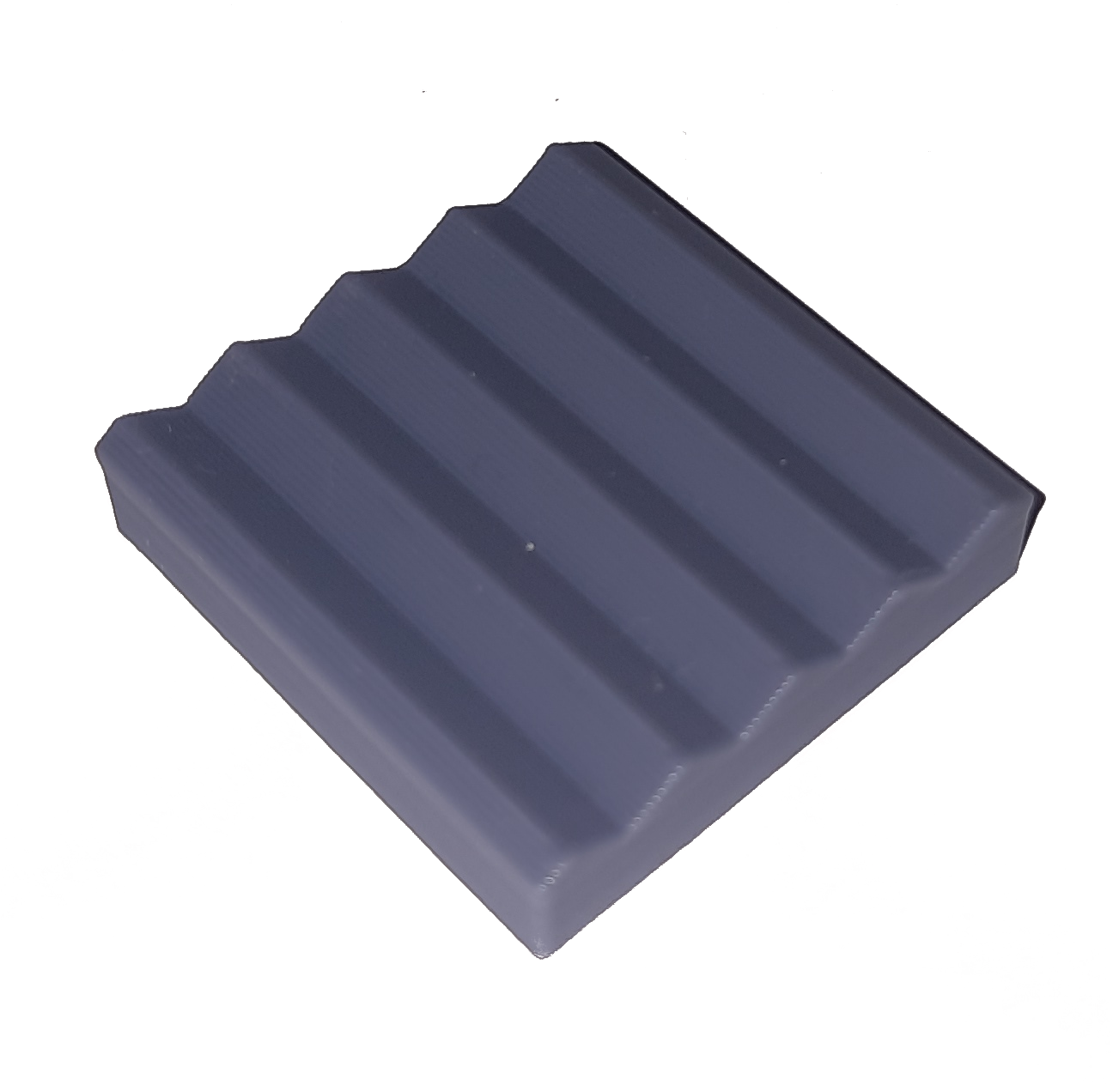}%
}
\subfloat[Texture 6]{%
  \includegraphics[clip,angle=0,width=0.33\columnwidth]{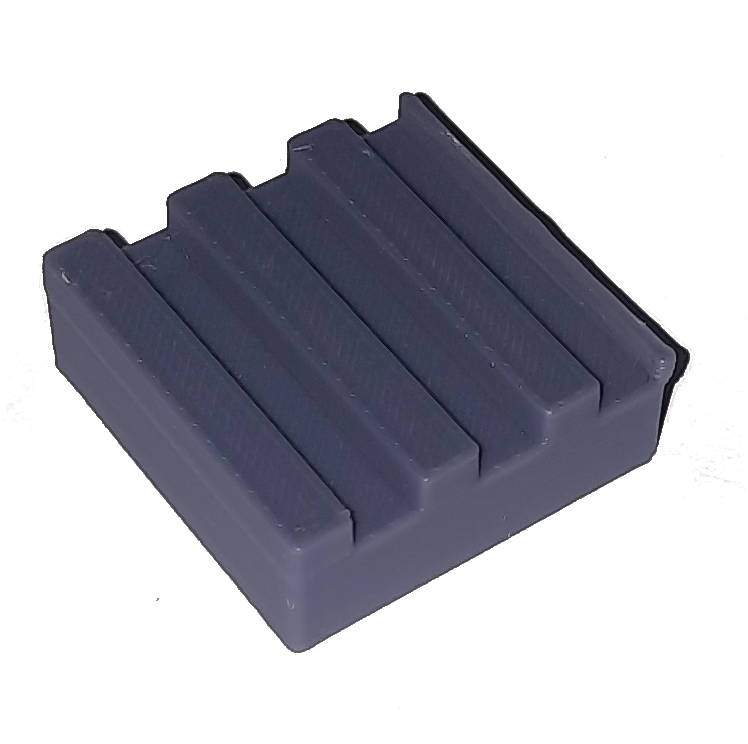}%
}

    \caption{The 3D printable textures printed in PLA- on a Bambu printer. }
    \label{fig:new-textures-printed}
\end{figure}

\subsection{Variability Across Printer and Filaments}

% todo show some varying quality of prints 
\begin{figure}[H]
    \centering
    \subfloat[Resin]{%
    \includegraphics[width=0.3\linewidth]{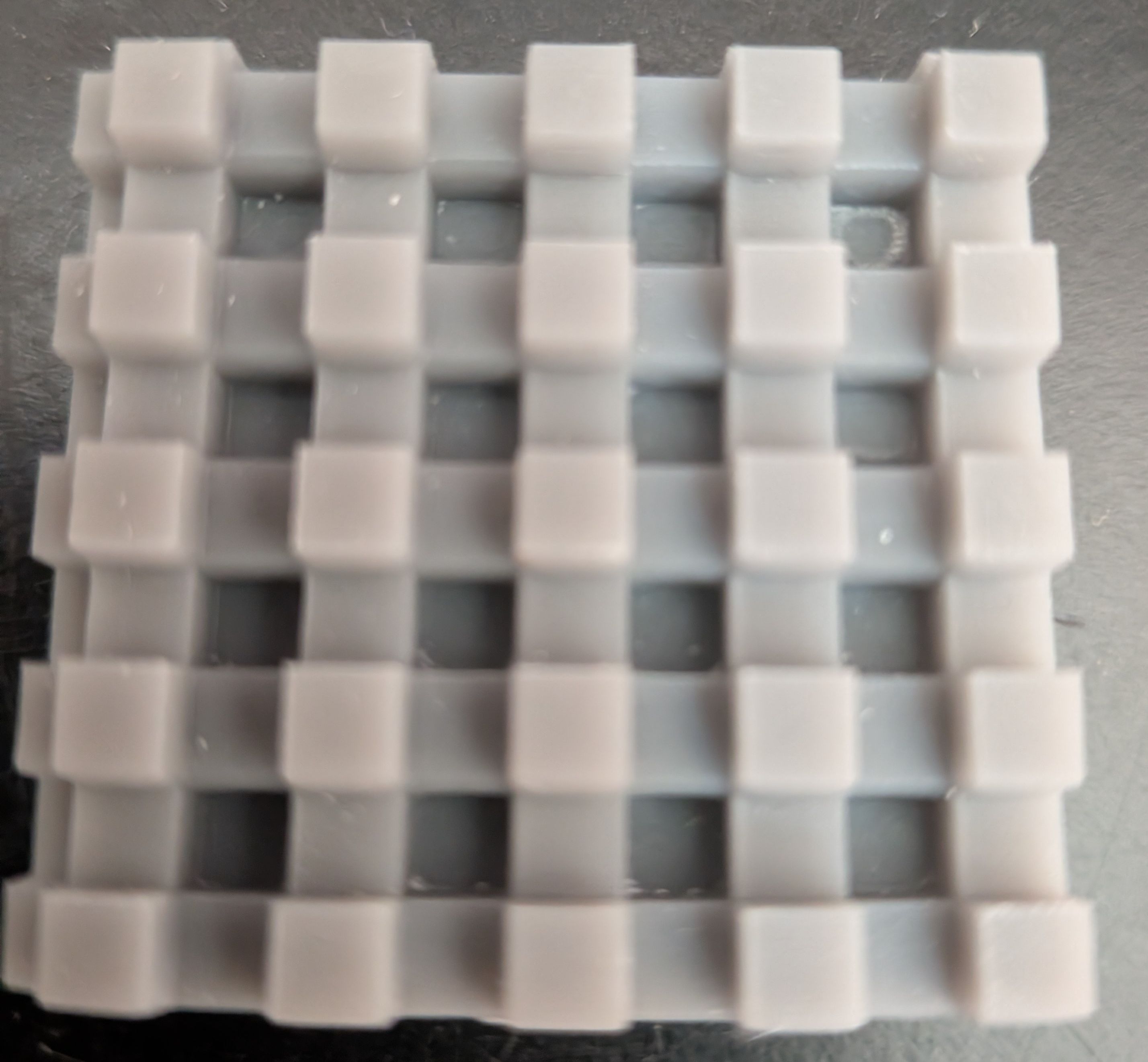}
    }
     \subfloat[Ender-3 PLA+]{%
    \includegraphics[width=0.3\linewidth]{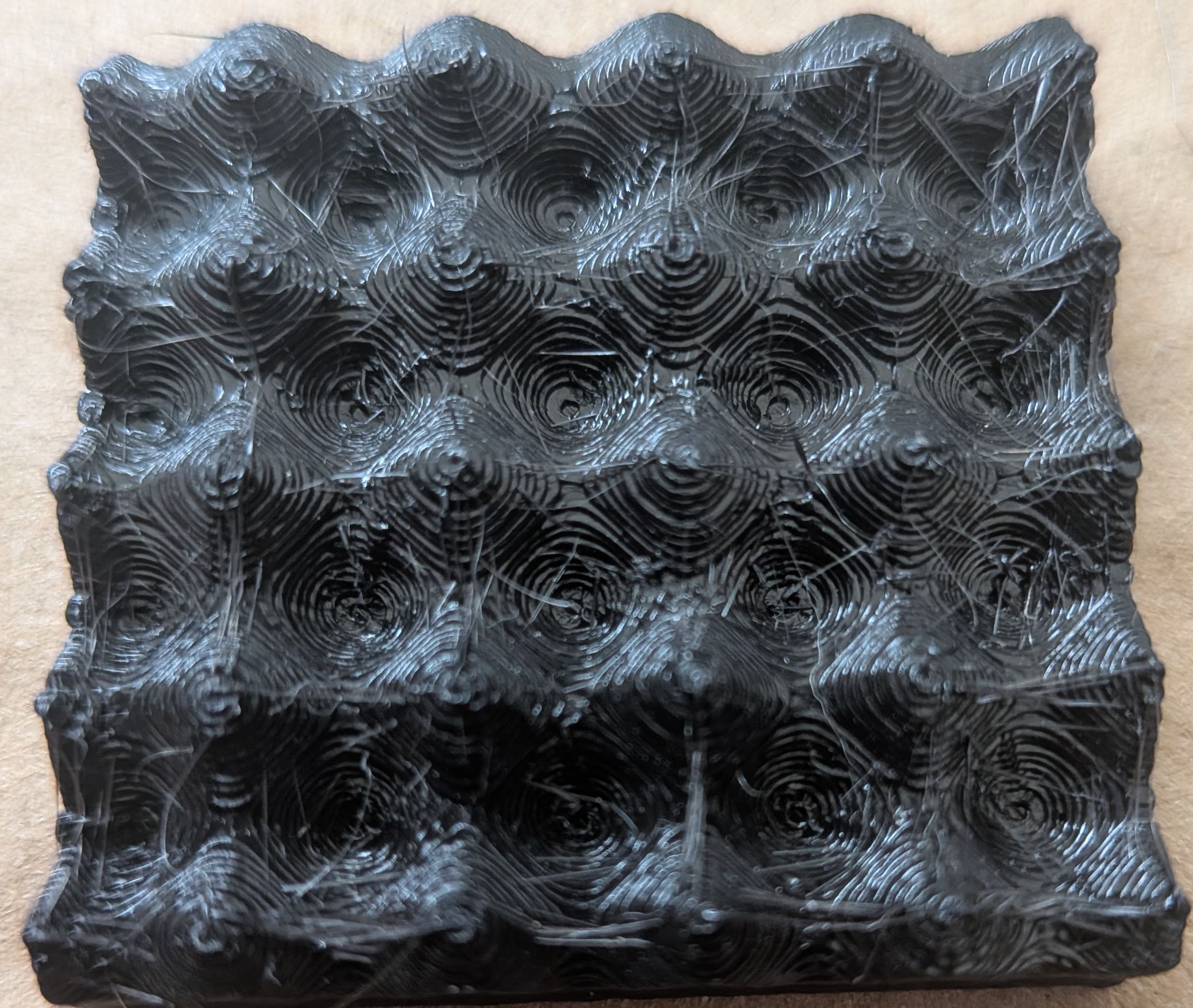}
    }
     \subfloat[Bambu PLA-]{%
    \includegraphics[width=0.3\linewidth]{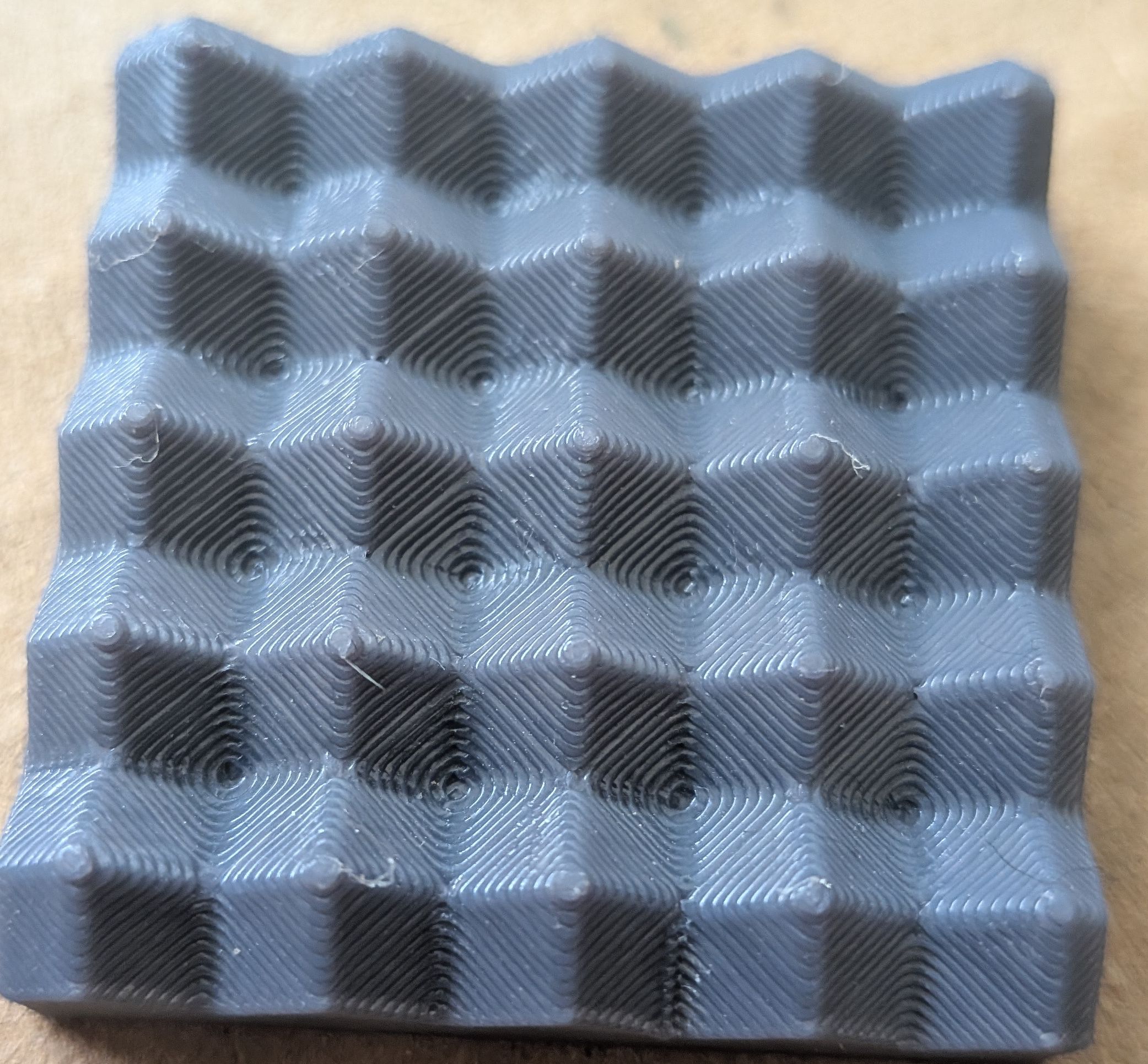}
    }
    \\
    \subfloat[Creality Ender-3 V3 SE PLA+]{%
    \includegraphics[width=0.3\linewidth]{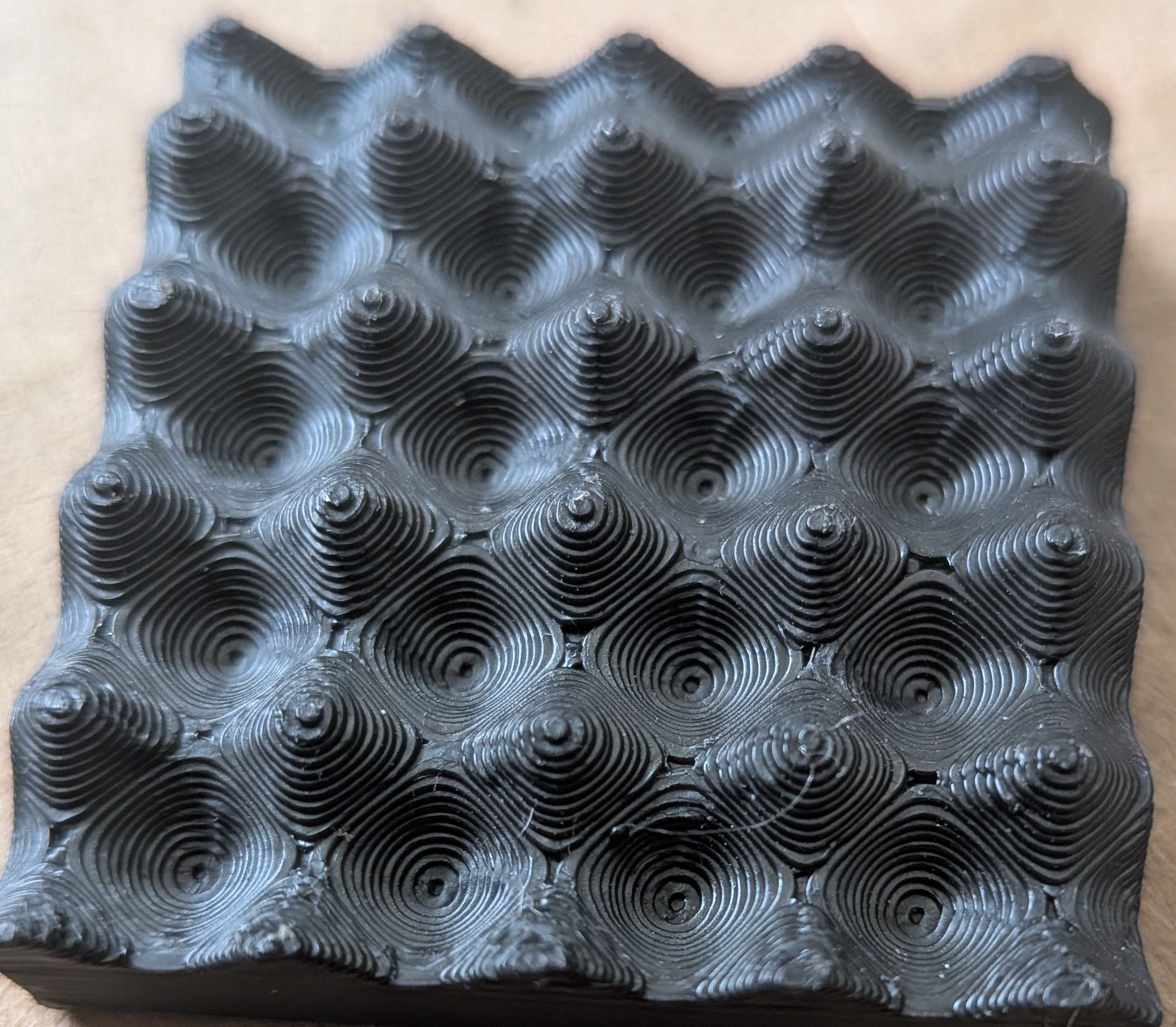}
    }
     \subfloat[Ender-3 PLA-]{%
    \includegraphics[width=0.3\linewidth]{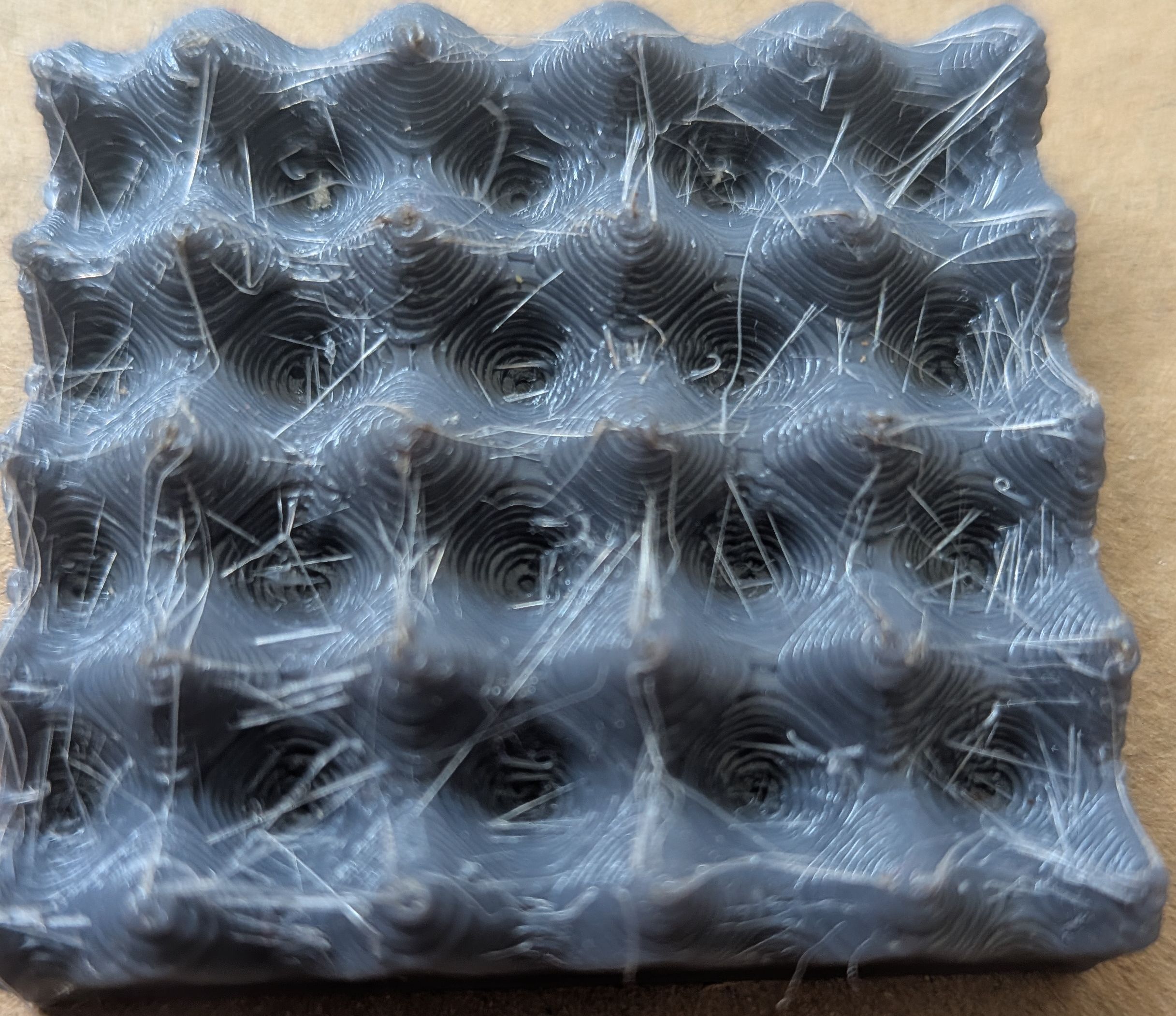}
    }
    \\
    \subfloat[\corr{Bambu and Ender-3 repeated prints}]{%
        \includegraphics[width=1\linewidth]{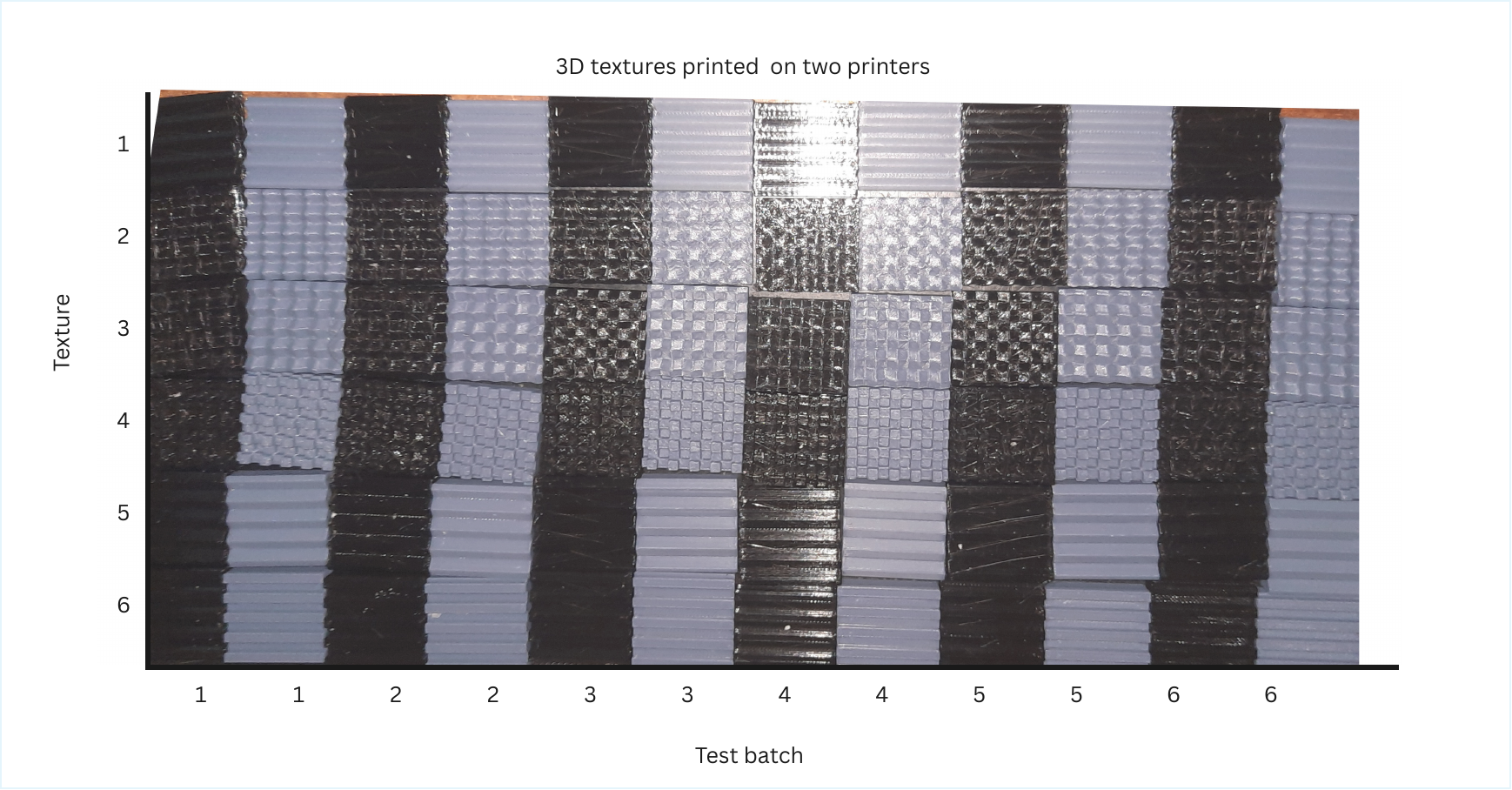}
    }
    \caption{Examples of the prints. The black filament (close up shown in \emph{(b) \& (d)}) shows the Ender-3 printer using PLA+, whereas \emph{(c)} shows the Bambu prints and \emph{(e)} Ender-3 both with PLA- which is grey. The cheaper printer (Ender-3) had a lot more stringing on all filaments used, and ridges were more visible. This undoubtedly led to higher variance. It is clear that there are no stringing effects on the resin printer (\emph{a}), making it a high quality print. In \emph{(f)} we show an example of the six textures printed five times on the Bambu with PLA-(grey) and Ender-3 with PLA+ (black).}
    \label{printbatch}
\end{figure}
%some with laces in and some not

%some where the points are not as pointy as they should be
To assess how much prints from different printers and filament combinations varied, we first used visual inspection to judge the quality of a print, particularly for "stringing". Stringing is when the printer’s nozzle leaks a small amount of molten filament while moving between separate parts or sections, leaving thin strands of plastic connecting them. If, from a human perspective, quality varies, this likely means the print quality has high variance. 

When we inspect the prints, some have more plastic webbing/string between blocks, especially with prints produced with lower-cost PLA (seen in Figure~\ref{printbatch}b\&d where there are thin plastic lines between the peaks), but these did not affect the surface of the block. More importantly, some of the peaks are more defined than other peaks on the same print. The sensor for dataset gathering comes from the top and presses down so stringing between peaks is unlikely to be picked up at this scale. However, differences in peak heights and well defined summits will be.

We next selected a subset of printers that spans the print-quality spectrum (as judged \corr{visually}) for more detailed evaluation: the highest-performing filament printer (Bambu Lab) paired with the lower-cost PLA- filament (henceforth referred to as `Bambu'), the lowest-performing filament printer (Ender-3) paired with PLA+ (henceforth referred to as `Ender-3'), and the resin printer (henceforth `Resin') as a consistently low-variance baseline. The Creality Ender-3 V3 SE was excluded from the extremes-spanning subset as we already had a higher end and lower end printer, showing no significant difference. This selection enables detailed comparison of print variability across different levels of printer capability and material quality. 

To assess the quality of prints, we generate five prints of each texture from each of these printer-filament combinations we first assess how much prints vary based on 3D scans of the texture blocks generated and then assess how much images from a commonly used Tactile sensor differ between repeats. Finally, to assess whether these differences are functionally important, we assess how well texture classification models perform on the different datasets.  

\subsection{3D scan}
\label{3dscanres}
To assess print variability, we made five prints of each of the six textures for each of the three selected printer-filament combinations. We measure the 3D structure of the blocks using an RGBD camera and assessed the difference between repeats as described in Section~\ref{scanning}. \corr{To account for scan error, we placed the same block in the scanner for each texture and scanned it 5 times for each of the six textures. We then compared this difference to the differences between repeats to see if the difference between prints is statistically different to measurement noise.}

\begin{table}[H]
\centering
    \begin{tabular}{|p{1.4cm}|p{1.4cm}|p{1.4cm}|p{1.4cm}|p{1.4cm}|p{1.4cm}|p{1.4cm}|}
        \hline
        \textbf{Printer Class} & \textbf{Texture 1} & \textbf{Texture 2} & \textbf{Texture 3} & \textbf{Texture 4} & \textbf{Texture 5} & \textbf{Texture 6} \\
        \hline
        \hline
        Ender-3 raw error
        & $4.21 \pm 1.41$ (n=20)
        & $3.63 \pm 1.11$ (n=20)
        & $3.35 \pm 0.56$ (n=20)
        & $4.11 \pm 0.88$ (n=20)
        & $4.41 \pm 1.26$ (n=20)
        & $6.17 \pm 2.41$ (n=20) \\
        \hline
        Ender-3 method error
        & $1.96 \pm 0.36$ (n=20)
        & $2.79 \pm 1.56$ (n=20)
        & $1.87 \pm 0.27$ (n=20)
        & $1.90 \pm 0.30$ (n=20)
        & $2.06 \pm 0.43$ (n=20)
        & $2.18 \pm 0.56$ (n=20) \\
        \hline
        Ender-3 T-test p-value
        & <1$\times 10^{-3}$
        & 7.51$\times 10^{-3}$
        & 9.99$\times 10^{-1}$
        & <1$\times 10^{-3}$
        & <1$\times 10^{-3}$
        & <1$\times 10^{-3}$ \\
        \hline
        \hline
        Bambu raw error
        & $2.52 \pm 0.45$ (n=20)
        & $2.39 \pm 0.19$ (n=20)
        & $2.99 \pm 0.69$ (n=20)
        & $2.19 \pm 0.34$ (n=20)
        & $2.54 \pm 0.40$ (n=20)
        & $2.02 \pm 0.30$ (n=20) \\
        \hline
        Bambu method error
        & $1.89 \pm 0.22$ (n=20)
        & $1.67 \pm 0.35$ (n=20)
        & $1.82 \pm 0.82$ (n=20)
        & $1.51 \pm 0.28$ (n=20)
        & $1.83 \pm 0.20$ (n=20)
        & $1.61 \pm 0.34$ (n=20) \\
        \hline
        Bambu T-test p-value
        & <1$\times 10^{-3}$
        & <1$\times 10^{-3}$
        & <1$\times 10^{-3}$
        & <1$\times 10^{-3}$
        & <1$\times 10^{-3}$
        & <1$\times 10^{-3}$ \\
        \hline
        \hline
        Resin raw error
        & $2.78 \pm 1.46$ (n=20)
        & $4.10 \pm 1.27$ (n=20)
        & $3.01 \pm 1.47$ (n=20)
        & $2.27 \pm 0.30$ (n=20)
        & $2.42 \pm 0.12$ (n=20)
        & $2.96 \pm 0.63$ (n=20) \\
        \hline
        Resin method error
        & $2.01 \pm 0.25$ (n=20)
        & $2.04 \pm 0.95$ (n=20)
        & $2.05 \pm 0.51$ (n=20)
        & $2.23 \pm 0.57$ (n=20)
        & $2.21 \pm 0.17$ (n=20)
        & $1.84 \pm 0.90$ (n=20) \\
        \hline
        Resin T-test p-value
        & 1.67$\times 10^{-2}$
        & <1$\times 10^{-3}$
        & 0.81$\times 10^{-3}$
        & 6.86$\times 10^{-3}$
        & <1$\times 10^{-3}$
        & <1$\times 10^{-3}$ \\
        \hline
        % \hline
        % \textbf{Ender-3 PLA+} & 17.51 & 18.4 & 51.26 & 65.59 & 63.88 & 49.26 \\ 
        % \hline
        % \textbf{Ender-3 PLA-} & 62.75 & 28.12 & 43.70 & 58.35 & 51 & 114.54 \\ 
        % \hline
    \end{tabular}
    \caption{Results for the difference calculations specified in Section~\ref{scanning} for the Ender-3 using PLA+, Bambu Lab using PLA-, and FormLabs Resin printer using grey V4.1. For each printer, we show mean $\pm$  standard deviation of the raw error in one row, and method error in the subsequent row, in units of $10^{-1}$mm. Both have the number of samples $n$ stated. This table includes a row from T-tests between the error and method error, to evaluate whether the difference between prints is significantly greater than the error from the method.}
    \label{tab:printer_errors_scan}
\end{table}

The average point cloud deviations for each manufacturing method are presented in Table~\ref{tab:printer_errors_scan}. The raw distance errors are the differences between scans of repeated prints of a texture after Iterative Closest Point (ICP) alignment. The method error are differences between scans of the same block, providing an empirical baseline of the overall repeatability error. As can be seen, the Ender-3  is the worst performing overall of our sample of printers with higher difference between print batches than the Bambu and Resin (aside from Texture 2 for the resin printer). As the method errors are broadly similar between printers, these differences are due to print quality.  Aside from Texture 2 and 6 to a lesser extent, the Bambu and Resin have similar differences with the Bambu superior on 4 textures, and Resin slightly better on 2. However, differences between repeated prints lie near the noise floor (i.e. method error differences) so it is unclear how significant these changes are. 
% Additionally, the Resin printer shows fewer repeats because lighting variation, which for most trials did not impact data gathering, affected one resin raw scan for texture 4, and XX method scans for Ender-3 texture 3. These outliers were removed from the sample.

Because of the difficulties of disambiguating differences from noise, and to assess how much these differences affect the behaviour of a typical sensor, in the next section we assess how much the response of a Tactile sensor varies across repeated texture prints.

\subsubsection{Tactile Image}
\label{imageres}
\begin{figure}[H]
    \centering
    \includegraphics[width=1\linewidth]{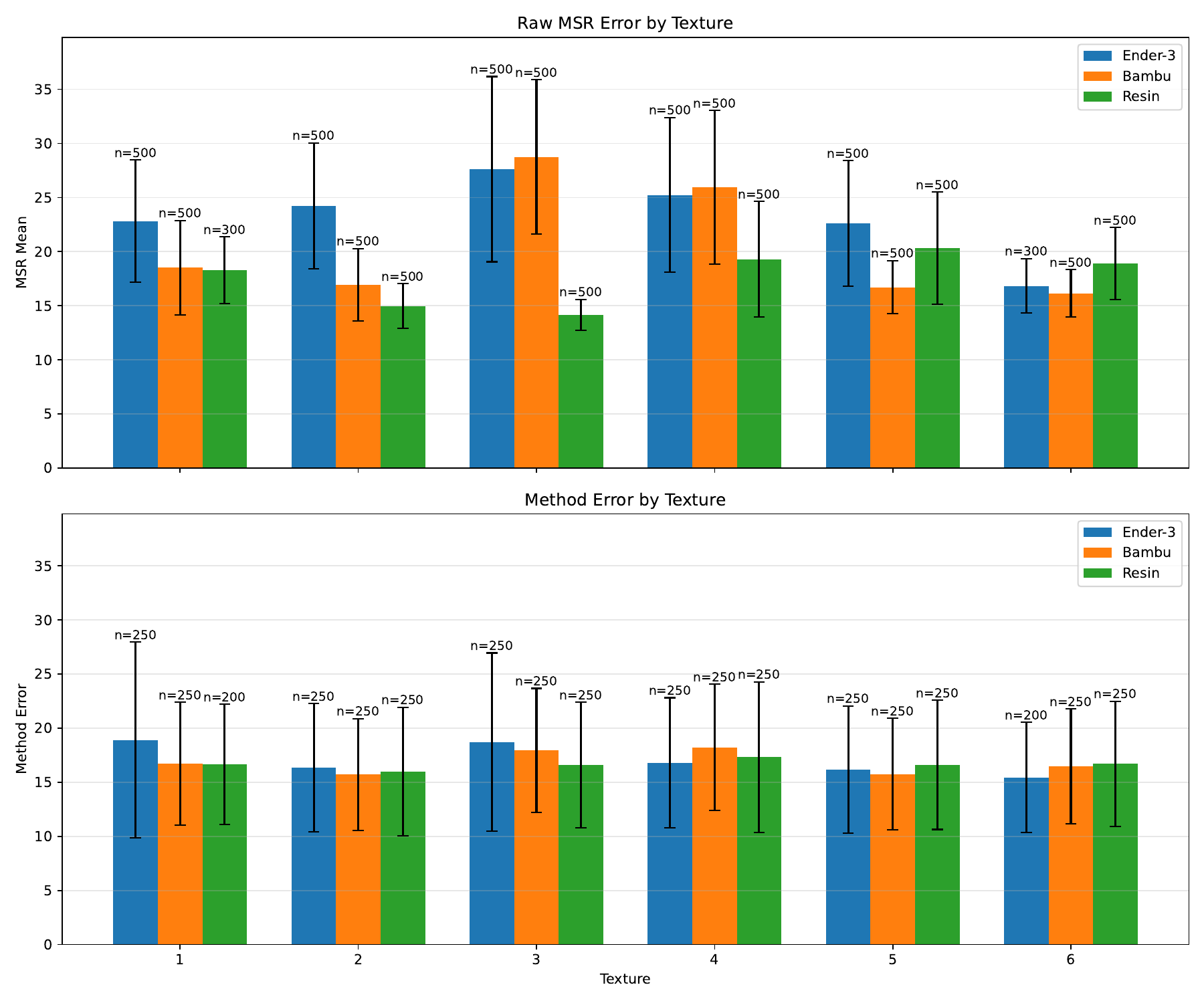}
    \caption{The top chart shows averaged image difference for three different printers across each texture with error bars showing the standard deviations of pixel difference. Data is collected from n=300-500 images for each texture (with the amount indicated above the error bar). This table includes data for the Ender-3 using PLA+, Bambu Lab using PLA- and FormLabs Resin printer using grey V4.1. The bottom chart shows the mean and standard deviation of the method error, where repeat presses of the same block are compared. This is made up of samples of n=200-250, with n shown above the error bar.  }
    \label{tab:variance_pattern}
\end{figure}

To assess variability between Tactile Images, we collected TacTip readings from 25 positions shown in Figure~\ref{fig:griddes}, taking 2 readings per position, across each repeat texture. For each texture and printer/filament combination, we then calculated the mean pixel-level difference between repeat trials for each of the 25 positions between unique combinations of repeats as in equation \ref{MSE_DIFF}. We refer to this as the raw error. To assess a method error, instead of calculating the differences between repeats, we calculated the difference between each of the two presses on the same block for each position. The results for each texture and grouped by printer-filament combination are shown in Figure~\ref{tab:variance_pattern}. As seen in the differences from 3D scans, while all printers had similar method errors, raw differences varied by texture and printer. Of the filament printers, the Bambu printer had a lower difference between repeated prints for four of the six textures, though had a slightly higher difference for textures 3 and 4. The resin printer had the most consistent difference across textures and the lowest overall difference for textures 1 to 4. Overall, however, as with the 3D measurements, there was no outright `winner' but a more general pattern of the Ender-3 being slightly worse than the other two printers.

%We calculate the average of variance between images of the same position with the same pattern across various trials, filaments and printers. The constant variable is the position of the sensor over the pattern block. This meant that each variance comparison was made up of 9 snapshots, averaged over 50 positions.

\begin{figure}[H]
    \centering
    \subfloat[Ender-3 ]{%
    \includegraphics[width=0.5\linewidth]{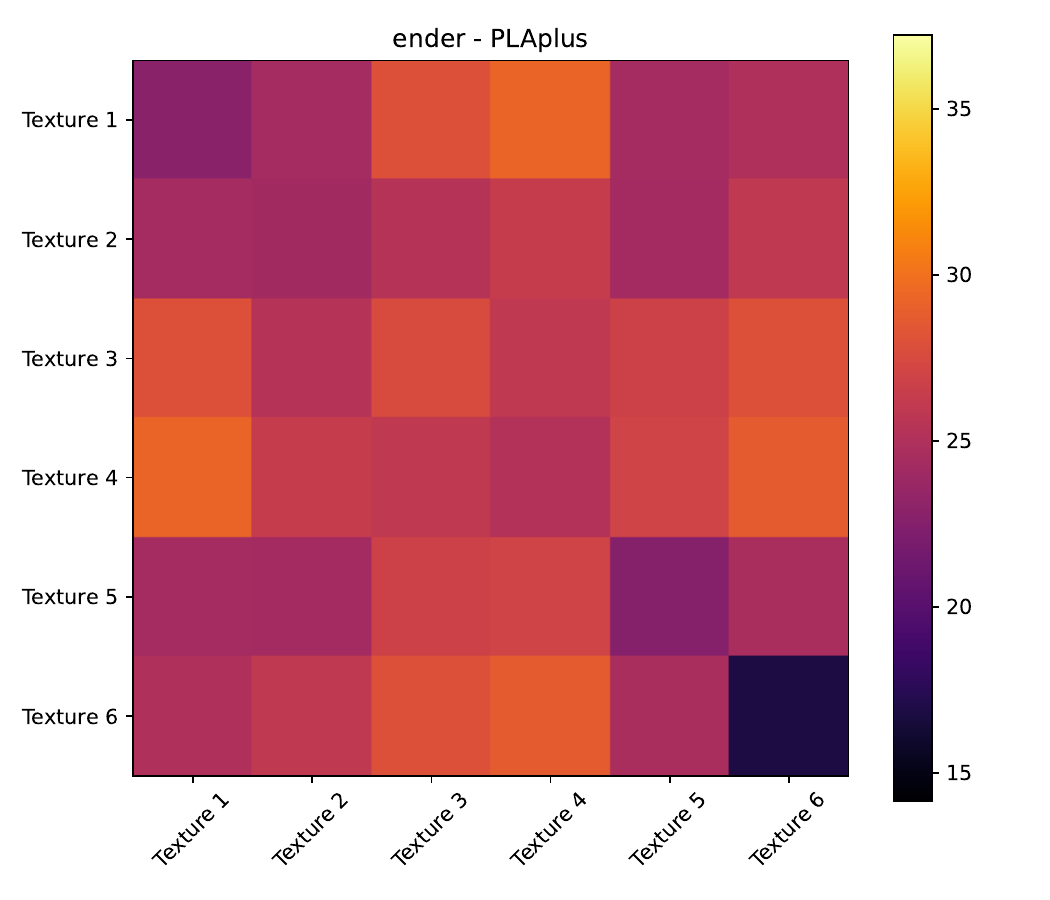}
    } % 352/600 =58%
    \\
    \subfloat[Bambu]{%
    \includegraphics[width=0.5\linewidth]{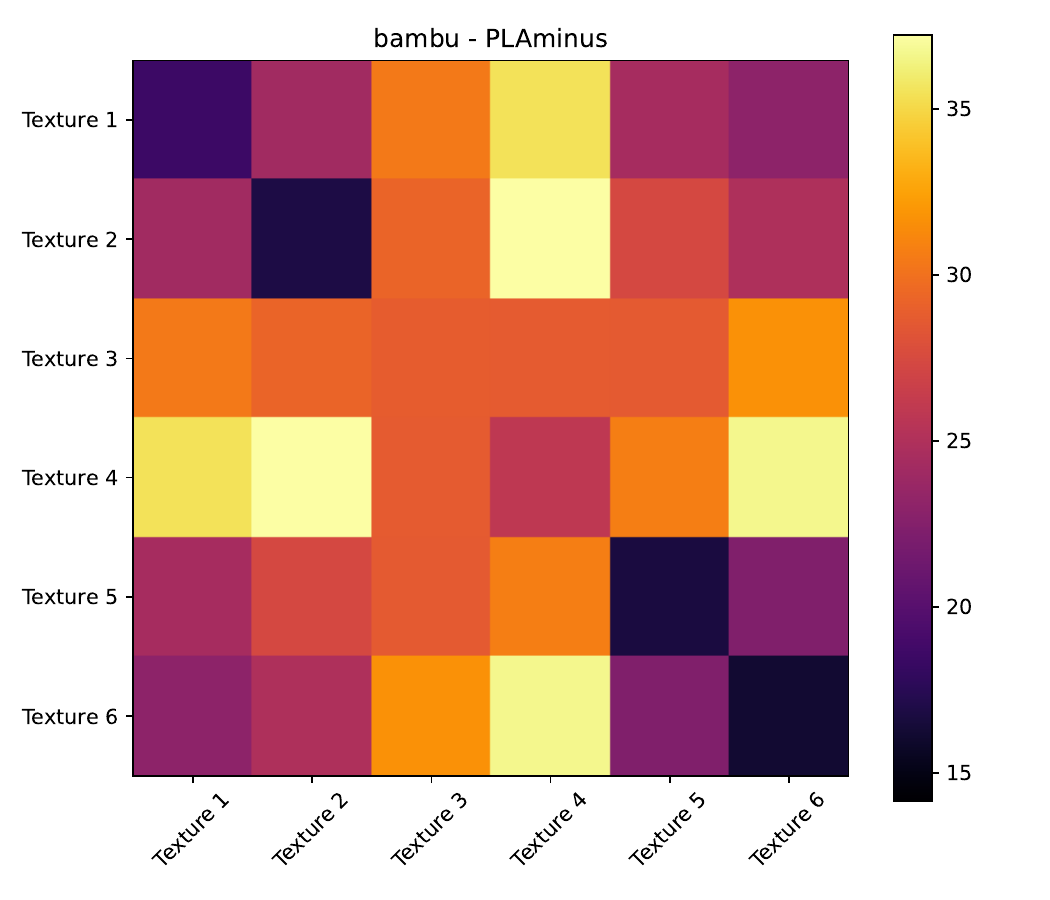}
    } %453/596 = 76% 
    \subfloat[Resin]{%
    \includegraphics[width=0.5\linewidth]{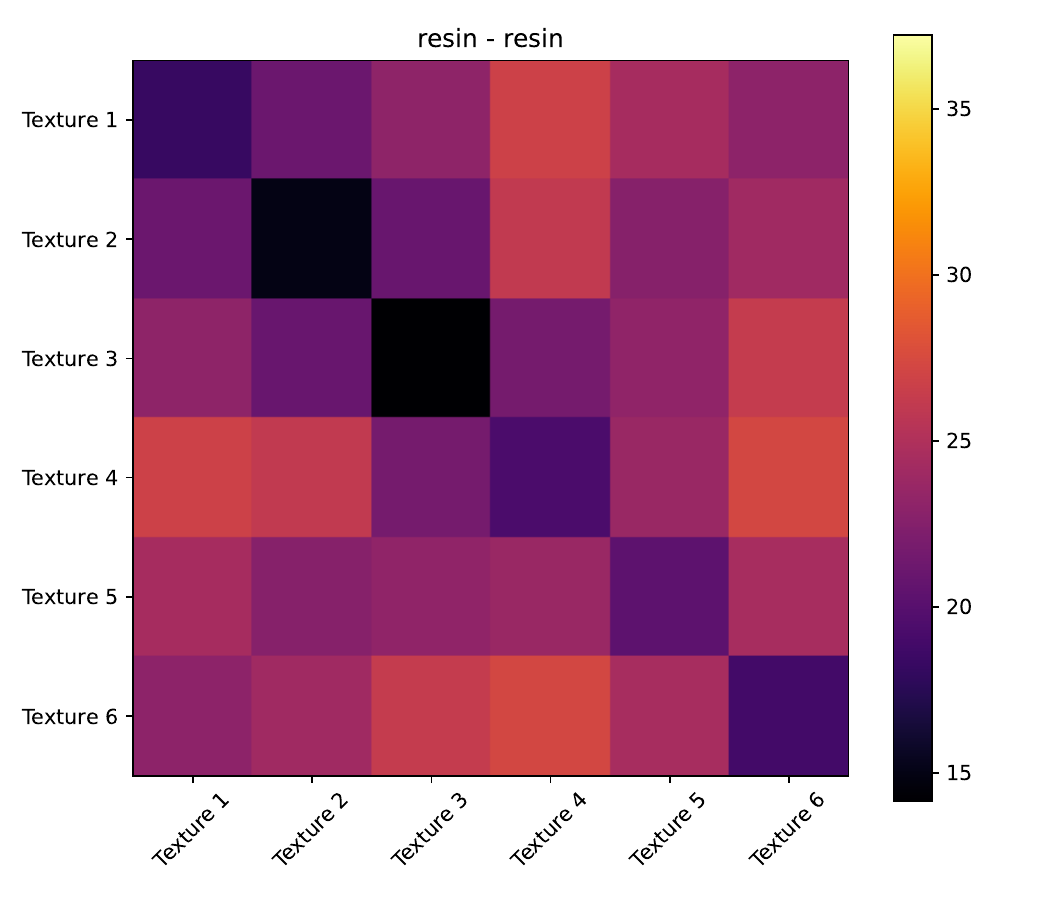}
    }
    
    \caption{The mean difference between pixels for each texture, compared across all of the textures. This table includes data for the Ender-3 using PLA+, Bambu Lab using PLA- and FormLabs Resin printer using grey V4.1. }
    \label{variance_matrix}
\end{figure}

As we are going to use this data for classification, we also examined the average differences between different textures (Figure~\ref{variance_matrix}), as this can tell us whether the differences in repeated prints would mask differences between textures. The Ender-3 has the most overlap between texture images (as judged by the difference scores), reflecting the  high differences between print batches of the same textures (aside from texture 6). The Bambu has a clearer trend in lower differences for the same texture, though textures 3 and 4 have high differences throughout, with texture 3 looking especially complex. These features would be harder to classify. The resin printer has the clearest picture overall with the diagonal elements the lowest and then rising differences off-diagonal.  The Bambu additionally has a lot of crossover between textures 1 and 2, and textures 5, and 6, where we see two blocks of low difference. On closer inspection, both the Ender-3 and Resin printers show similar, if less obvious, lower difference for these patterns. This reflects the similarities between waves 1 and 2 and 5 and 6 from a Tactile sensor's perspective (which emphasises the higher points of the patterns).

\subsection{Effect of printer variability on texture classification}
In the previous section we saw that different printers differ in print variability. We next wanted to know whether this variance has an effect on classification. To do this we trained classifiers to classify textures from different printer and filament combinations. \corr{To mitigate time-varying sensor artifacts, such as marker tracking noise or transient illumination fluctuations, two independent readings were recorded at each relative texture position. Furthermore, systematic sensor behaviours (e.g., elastomer deformation and hysteresis) were kept constant across all trials by utilising a single TacTip sensor profile under identical robotic displacement constraints, ensuring that variations in the signal directly reflect physical manufacturing defects.} To make sure our results were not due to a particular classifier, we trained both a random forest classifier and a feedforward artificial neural network (ANN) (models chosen following our previous work on tactile texture classification~\cite{shepherd2025texture}). A Random Forest Classifier was trained using the default scikit-learn hyperparameters. The model comprised 50 decision trees, using the Gini impurity criterion for splitting. Trees were allowed to grow without a predefined maximum depth, while  $\sqrt{n}$ features were considered at each split. Bootstrap sampling was enabled, with the remaining parameters retained at their scikit-learn default values. This was in line with previous experiments~\cite{shepherd2025texture}. The ANN used the same setup as existing TacTip classification~\cite{shepherd2025texture} namely a single hidden layer of size 50 and an output layer of 6. To train the ANN we used cross entropy loss with the Adam optimizer at a learning rate of 0.001. Training took place over 1500 epochs. These models were trained from randomly split data, and random initialisation states of the models, therefore our results shown are averaged over 20 trials. 

\begin{figure}[t]
    \centering
    \includegraphics[width=0.8\linewidth]{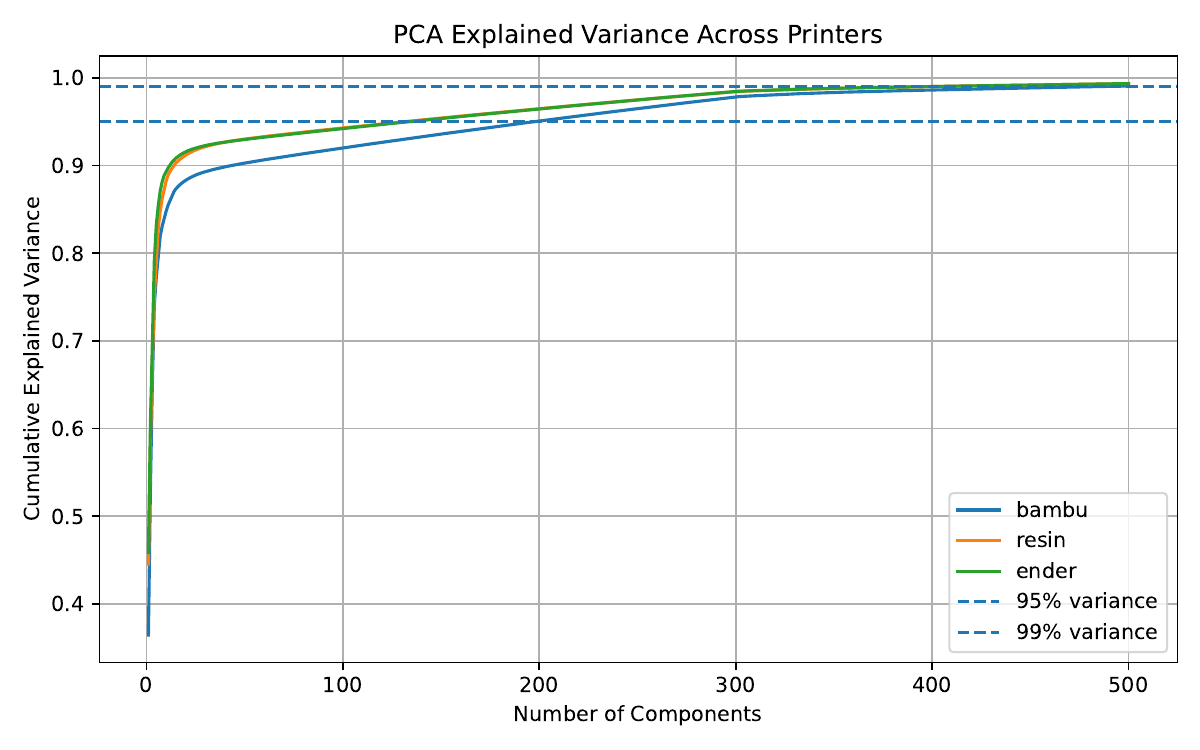}
    \caption{\corr{Cumulative explained variance of principal component analysis (PCA) for tactile images collected from the Bambu, resin, and Ender-3 printers. The cumulative variance is plotted against the number of principal components required to represent each dataset. Dashed horizontal lines indicate the 95\% and 99\% explained variance thresholds. Differences between the curves reflect the intrinsic variability of tactile features produced by each printing process, with datasets requiring more principal components exhibiting greater variation in surface characteristics. } }
    \label{PCA_EXPLAINED}
\end{figure}

%2 x 25 x 6 = 300 x 5 = 1500 x 3  = 4500
The dataset consisted of all 5 repeats of each of the six textures produced using all printer–filament selections. As there are two presses at 25 positions this resulted in  2 x 25 x 5 (repeats) x 6 (textures) x 3 (printer-filaments) = 4500 images. Removing the two corrupted samples left a dataset of 4300 images. To reduce memory requirements, all images were converted to greyscale and downsampled to 40\% of their original resolution, a preprocessing step previously shown not to affect classification performance~\cite{shepherd2025texture}. After preliminary investigation, we also decided to use PCA to convert the dataset to a smaller dimensionality thus removing noise and highlighting useful features for the classifiers. \corr{Analysis of the amount of variance explained by PCA components (Figure~\ref{PCA_EXPLAINED}) shows that the amount of variance explained plateaus around 300 components, and thus was chosen to keep the data dimensionality low without dropping meaning in the data.}  PCA was fitted to \corr{the training split of the dataset} to convert the images to 300 principal components (chosen based on preliminary results), \corr{then the test split was transformed using the fitted PCA}. The principal components were then used as input to both the Random Forest Classifier and the ANN. 

We also applied augmentation to the training datasets to account for LED noise and global lighting variation. Specifically, for each image we added Gaussian noise with mean 0 and standard deviation 5 to each pixel, making one new image (`Noise' augmentation). We also added/subtracted 10 to each pixel to `vary' the lighting, resulting in 2 new images (`Brightness' augmentation). To see if these augmentations were useful we ran a preliminary test using only the RFC and with only 25 PCA components. The initial test used 25 components because it was faster and results showed good accuracy; however, upon further experimentation we found that 300 would lead to higher accuracy and used this for later experiments. Because we wanted to assess whether the augmentation helped with variability across prints, we trained the RFC on all the data from batch 1 and tested it on batch 2, with the accuracy being the performance on the test set. \corr{The results of the ablation study are shown in Table~\ref{tab:augmentation_ablation_printers}}. Brightness did not help accuracy, therefore we continued with noise injection only. 

\begin{table}[t]
\centering
\begin{tabular}{llc}
\hline
\textbf{Printer} & \textbf{Experiment} & \textbf{Accuracy (\%)} \\
\hline

\multicolumn{3}{c}{\textbf{Ender-3}} \\
\hline
& No augmentation & 54.15\% $\pm$ 0.03 \\
& Brightness & 54.76\% $\pm$ 0.03 \\
& Noise & 76.52\% $\pm$ 0.03 \\
& All augmentations & 73.23\% $\pm$ 0.02\\
\hline
\multicolumn{3}{c}{\textbf{Bambu}} \\

\hline
& No augmentation & 89.52\% $\pm$  0.02\\
& Brightness & 89.5\% $\pm$ 0.02 \\
& Noise & 91.65\% $\pm$ 0.01 \\
& All augmentations & 90.24 \% $\pm$   0.02\\
\hline
\multicolumn{3}{c}{\textbf{Resin}} \\
\hline
& No augmentation & 80.40\% $\pm$ 0.03\\
& Brightness & 79.99\% $\pm$ 0.02 \\
& Noise & 94.98\% $\pm$ 0.01 \\
& All augmentations & 94.50\% $\pm$ 0.01 \\
\hline

\end{tabular}

\caption{\corr{Data augmentation ablation study across different 3D printers, using 25 components with PCA. This data is for a Random Forest Classifier showing mean and standard deviation over 20 trials. The classifier was trained on batch 1 and tested on batch 2.} This table includes data for the Ender-3 using PLA+, Bambu Lab using PLA- and FormLabs Resin printer using grey V4.1. }
\label{tab:augmentation_ablation_printers}
\end{table}

% Table~\ref{tab:3dclassifierresults} assesses global generalisation via a Leave-One-Print-Out split across all batches produced by that printer model. 

%The random forest Classifier , as it has also been seen to work well on these tasks~\cite{shepherd2025texture}.
%todo show matrix

%The performance degradation observed for the Bambu system in Table~\ref{tab:3dclassifierresults} highlights how micro-scale manufacturing variability across distinct print nodes degrades classifier performance, which has not been observed to the same reduction for the Resin or Ender-3 classification. Looking back to Figure~\ref{variance_matrix} shows a higher difference confusion between the Bambu textures than the Resin. 

%todo talk about PCA and how far off predictions are 
After establishing which augmentation worked well we then tested both ANN and RFC classifiers with the full dataset and 300 PCA components. To assess the impact of manufacturing repeatability, each classifier was trained/tested using a single batch of prints from a specific printer-filament configuration, but then additionally evaluated by testing the trained model on the other 4 datasets, which are completely unseen to the classifier training. Each batch comprises 50 images per texture so With noise augmentation doubling the number of imaages, gives 600 images per printer-filament batch. We used a 70\% training/test split. This resulted in 420 training samples and 180 same-batch test samples. After training, we tested the models on the remaining 4 batches of data (with data augmentation, making X images in total) with the accuracy referred to as the unseen print accuracy. Results are shown across 20 runs in Table~\ref{tab:3dclassifierresults} where we report classifier performance within-batch (train and test accuracy - the latter being the accuracy on the test set) and cross-batch generalisation (unseen print accuracy). 
%An augmentation ablation study (Table~\ref{tab:augmentation_ablation_printers}) showed that noise augmentation provided the most robust performance and was therefore adopted for the remainder of the study. 
%We evaluated the Ender-3 PLA+, Bambu PLA- and resin configurations using larger datasets and the same cross-batch evaluation methodology using a 20\% test split. 

For the RFC, train and test results were near perfect, meaning it was hard to distinguish between printers. However, as can be seen in the third column of Table~\ref{tab:3dclassifierresults}, the resin printer achieved the best performance on unseen print batches followed by the Bambu with the worst results coming from the Ender-3. These results reflect the differences in 3D scans and tactile images and so we assume that the increased performance for the Resin is due to the greater fidelity of repeated prints. While classifiers might be partly exploiting print-specific manufacturing artifacts, such as stringing residue or layer-line patterns, rather than the intended texture geometry alone, particularly for the lower-fidelity FDM prints, the fact that the ordering of results reflects the ordering seen in differences between tactile images suggests that repeatability is also a factor. We return to this point in the Conclusion. The results for the ANN were worse than the RFC for all cases (aside from the unseen accuracy on the Ender-3). However, the ordering of the results - with the Ender-3 much below the two other printers on unseen accuracy, again reflects the metrological data, showing that our results are not due to a particularity of the RFC. 

Overall the Bambu and resin printers' high cross-batch accuracy highlight the high material and geometric consistency captured by the tactile sensor. 

\begin{table}[H]
    \begin{tabular}{c|c|c|c}
    \textbf{Model} & \textbf{Train accuracy} & \textbf{Test accuracy} & \textbf{Unseen print accuracy}\\
    \hline
    RFC Ender-3 & 100.00\% $\pm$ 0.00 & 98.66\% $\pm$ 0.94 & 42.66 \% $\pm$ 2.25 \\
    ANN Ender-3 & 93.28\% $\pm$ 0.06 & 91.79\% $\pm$ 0.07 & 42.87\% $\pm$ 0.03\\
    RFC Bambu & 100.00\% $\pm$ 0.00 & 99.63\% $\pm$ 0.75 & 73.75\% $\pm$ 1.09\\
    ANN Bambu & 94.9\%$\pm$  0.07 & 92.20\% $\pm$ 0.10 & 57.46 \% $\pm$ 0.05\\
    RFC Resin & 100.00 \% $\pm$ 0.00 & 97.89\% $\pm$ 1.99 & 80.27\% $\pm$ 1.51\\
    ANN Resin& 92.28\% $\pm$ 0.06 & 89.71\% $\pm$ 0.07 &  56.35 \% $\pm$ 0.05\\
    \end{tabular}
    \caption{Model classification performance when trained on noise-only augmentation with PCA of 300 components. Training is done on 70\% of one batch with testing on 30\% of the same batch. Train/Test accuracy are the final accuracy on these batches. The 'Unseen print accuracy' columns show the model's performance when evaluated on the larger set of print batches held out entirely from training, utilising the same printer and filament configuration. We then average the results of the performance over 20 runs.} 
    \label{tab:3dclassifierresults}
\end{table}

% Our metrological analysis in Section 3, these high-fidelity micro-topographies sit below the statistical resolution limit of non-contact optical hardware ($p > 0.05$). The model's stable cross-batch accuracy on the resin platform confirms that the contact-based tactile pipeline successfully extracts invariant structural signatures embedded deep within this manufacturing noise floor, whereas the larger cross-batch drift on the Bambu system reveals a higher susceptibility to node-specific artifacts.

\begin{figure}[ht]
    \centering
        
    \subfloat[Ender-3 ]{%
    \includegraphics[width=0.5\linewidth]{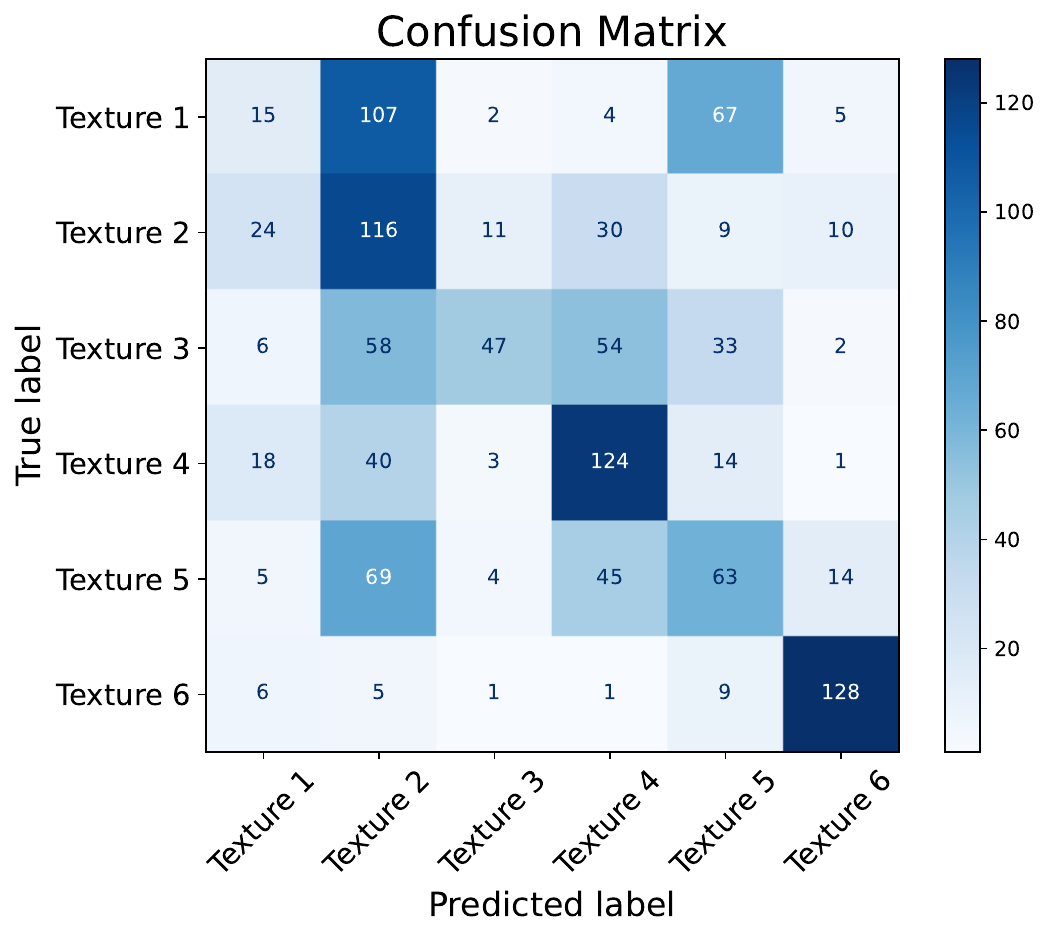}
    } % 352/600 =58%
    \\
    \subfloat[Bambu]{%
    \includegraphics[width=0.5\linewidth]{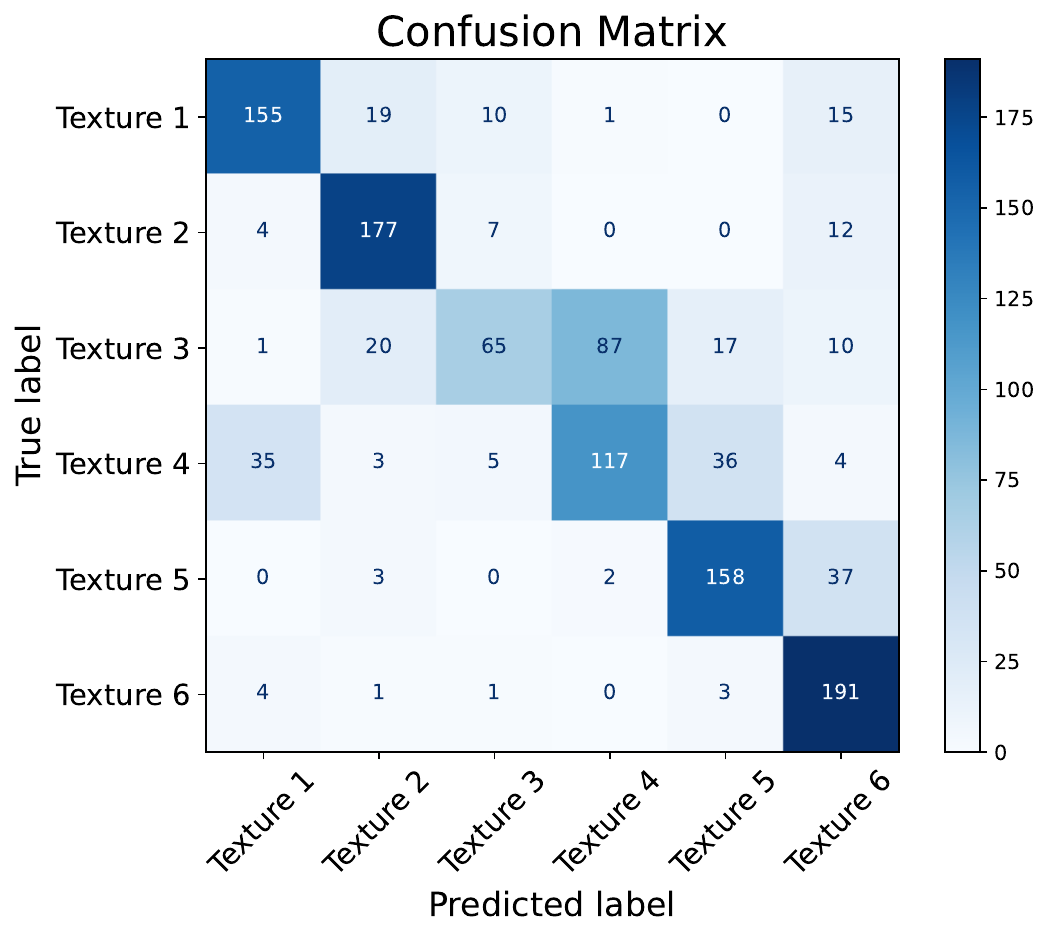}
    } %453/596 = 76% 
    \subfloat[Resin]{%
    \includegraphics[width=0.5\linewidth]{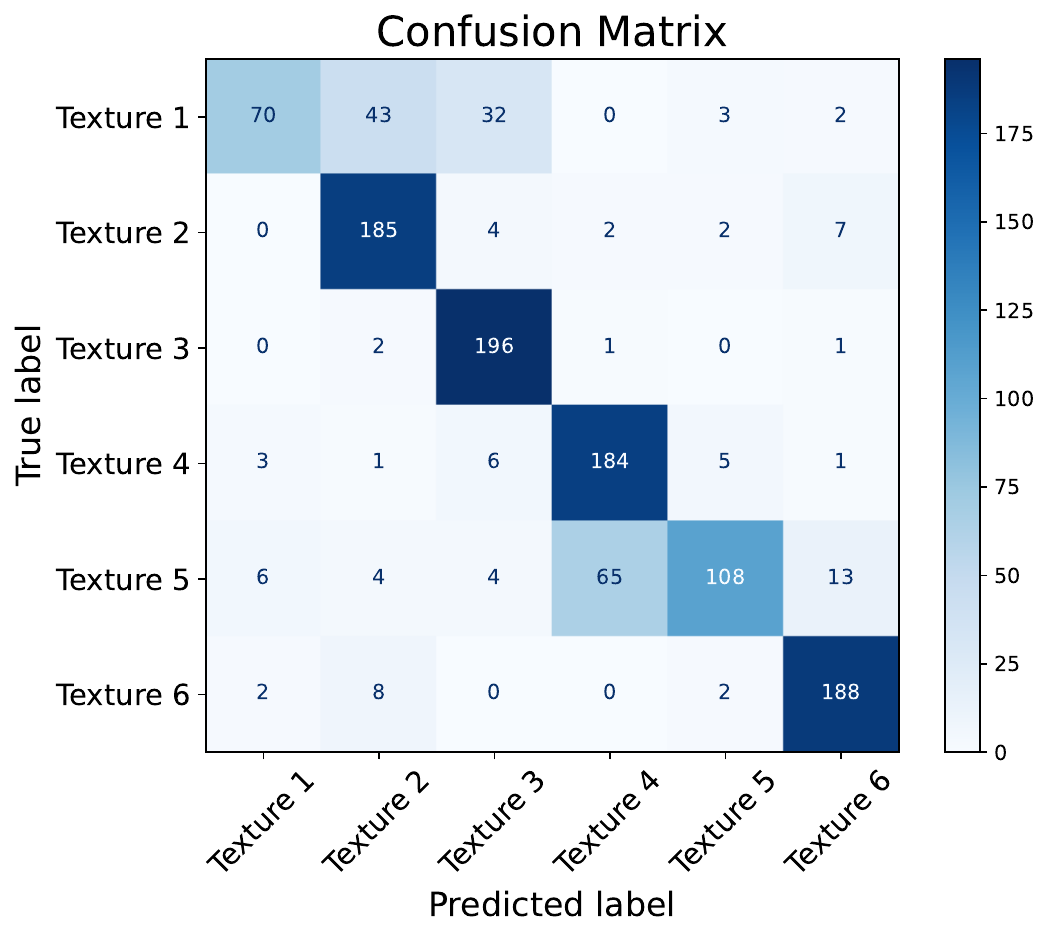}
    }

    \caption{\corr{Confusion matrix for the random forest classifier trained on the \emph{a}: Bambu dataset, \emph{b}: Resin dataset and \emph{c}: Ender-3. All evaluated on unseen other prints. Performance is high overall suggesting that the benchmark dataset contains texture features that are consistent across manufacturing conditions while remaining sufficiently challenging to assess classifier robustness and generalisation. }  }
    \label{fig:confusion3d}
\end{figure}

When examining the confusion matrices of the Random Forest Classifier across the different printing platforms (Figure~\ref{fig:confusion3d}), the majority of the tactile profiles are correctly classified. However the systematic errors reveal insights into the physical limits of the tactile sensing pipeline. 
%In particular, across printers, there is a tendency to misclassiffy Textures 1 and 2 reflecting the cross-texture tactile image differences of Figure~\ref{variance_matrix}.  

Incorrect classifications consistently revolve around specific geometric similarities. For instance, Texture 1 is frequently misclassified as 2 and 3. In the Ender-3 and Bambu confusion matrices 2 and 3 are misclassified which is likely because they are generated from similar spatial equations (see Figure~\ref{fig:new-textures-printed} for a visual reference). Crucially, a similar cross-talk occurs between Textures 1 and 5 within the Ender-3, and a large bias towards Texture 2. Interestingly, this is a one way relationship of misclassification showing learned biases between similar patterns. Similar relationships can be seen in Figure~\ref{variance_matrix} where there is structural similarity between specific textures. 

As illustrated in the Ender-3 confusion matrix (Figure~\ref{fig:confusion3d}c), the classification accuracy for Texture 1 degrades significantly to 7.5\%, with a notable 53.5\% of its data mass leaking directly into Texture 2. This performance drop confirms that the ultra-fine micro-geometrical traits of Texture 1 sit directly at the operational noise floor of the tactile sensor. On coarser extrusion systems like the Ender-3, these faint geometric signals are easily masked by macro-scale manufacturing noise (such as layer shifts and extrusion variations), causing the model to confuse it with the more pronounced features of Texture 2. 

Conversely, the much cleaner separation of Texture 1 on the high-fidelity resin platform demonstrates that when these baseline manufacturing artifacts are minimised, the tactile sensor's physical compliance can successfully extract the underlying texture signature. This contrast highlights that cross-print generalisation is heavily dictated by printer-node variability, where manufacturing inconsistencies act as an undesirable confounding variable rather than a robust tactile trait.

As a final note, consider the very high accuracies which result in quite clean confusion matrices compared to the tactile image differences seen in Figure~\ref{variance_matrix}. This shows that the contact-based tactile pipeline successfully extracts invariant structural signatures from the data.

\section{Conclusions}

This paper introduces a method for generating geometric patterned 3D printed \corr{rigid} textures that can be reliably reproduced across different research teams, enabling fair and consistent comparisons between sensors in texture classification tasks. Our investigations highlight that filament type and printer quality can influence the resulting texture fidelity. From the filament-fed printer models, the Bambu had less difference between print batches than the Ender-3, which also experienced more stringing and overall messier prints than the Bambu. Resin was visually best overall showing a lower difference between prints. 

Our metrology and classification results show the same pattern confirming that while lower-tier FDM hardware suffers from artifact-driven distortion, high-resolution resin printing significantly suppresses manufacturing variance, securing an unseen cross-print accuracy of 80.27\% using a Random Forest Classifier. Note that the geometric scan-based ranking (Bambu marginally ahead of Resin) and the tactile classification-based ranking (Resin marginally ahead of Bambu) disagree. Given that both printers' geometric deviations sit within, or close to, the optical scanner's noise floor (Section~\ref{3dscanres}), we treat this scan-based ordering as unreliable for distinguishing the two. 

Rather than invalidating the metrological pipeline, this highlights a physical justification for the tactile benchmarking methodology. It demonstrates that the cross-print structural variations introduced by modern additive manufacturing systems, even coarser filament-based FDM platforms, are so small that they sit around the resolution threshold of a 3D scan using a high-precision optical laser scanner. Thus this form of non-contact metrology does not reliably separate our high-fidelity print iterations. Further experiments using the TacTip image difference revealed similar patterns to the 3D metrology results. Conversely, the contact-based TacTip sensor successfully leverages localised physical surface compliance and skin deformation to map micro-geometrical traits. The Ender-3 typically had higher difference across print batches, whereas the Resin and Bambu remain similar. This high physical sensitivity allows the machine learning framework to extract consistent structural signatures embedded deep within the manufacturing noise floor, securing high cross-print validation accuracy on both Bambu and resin printers.

While the resin printer had the best classification performance on unseen data, and tended to lead to lower MSR within the image comparisons, the Bambu printer had similar results (and moreover, the 3D scan did not reveal evidence the Resin being better than the Bambu). 
% Among filament-based systems, higher-end printers such as the Bambu Lab performed competitively, while lower-cost systems such as the Ender-3 exhibited greater variability. 
Our evaluation suggests that high-resolution resin printing serves as a highly capable baseline. However, we acknowledge that the Bambu Lab paired with PLA provides a strong lower-cost alternative. Importantly, while we have only tested a limited number of printers, we have described repeatable methods that can be applied to compare prints from other printer and filament combinations to these.

\corr{Although the 3D scanning analysis quantifies the geometric fidelity of the printed textures, manufacturing imperfections remain a potential source of confounding. In particular, classifiers may exploit subtle print-induced artifacts in addition to the intended texture geometry, especially for lower-fidelity fabrication methods such as FDM printing. While our analysis characterises these differences in print quality, we cannot completely exclude the possibility that such artifacts contribute to classification performance. Future work should investigate the robustness of the benchmark across multiple printers and fabrication methods to further isolate texture-specific features from manufacturing-induced variation. The results indicate by using a resin printer, the potential issue of these sorts of noises could be largely eliminated. }

A limitation of this study is that evaluations were conducted using a single physical unit per printer model. Consequently, observed variations incorporate localised unit-level maintenance states alongside inherent model-level performance capabilities. Future cross-lab studies are required to isolate machine-specific wear from broad technology-class constraints.

While this work did not explore slicer-level optimisation or printer calibration, such directions offer clear opportunities to further reduce variability, particularly for lower-cost systems. In summary, this study provides a reproducible dataset, a methodology for producing such datasets, and a structured evaluation framework for quantifying and comparing the effects of 3D printing parameters on texture consistency, all of which can support future work in texture analysis.

\textbf{Acknowledgments.}
DS was funded by a be.AI Leverhulme Trust Doctoral Scholarship. We thank Nashil Sowaruth for 3D printing the resin models. 

\appendix
\setcounter{section}{0}
\label{appendixcode}
\section{Repository}
The code and experimental data are available at:
\url{https://github.com/shepai/3D-textures/}.
The naming conventions refereed to in Section~\ref{section:imageEval} correspond to:
 \begin{itemize}
     \item \emph{3DD1} - Found in repository folder Generator/objects/
     \item \emph{3DD2} - Found in repository file Generator/Generator.py
     \item \emph{3DD3} - Found in repository readme section "Data \& Experiments" part 1. 
 \end{itemize}
 
%3dd1 https://github.com/shepai/3D-textures/tree/main/Generator/objects
%3dd2 https://github.com/shepai/3D-textures/tree/main/Generator/Generator.py
%3dd3  readme

\setcounter{section}{1}
\label{appendix}
\section{Intel RealSense D405 Configuration}
\begin{table}[H]
\centering
\caption{Intel RealSense D405 configuration used in our experiments.}
\label{tab:d405_config}
\begin{tabular}{ll}
\hline
\textbf{Property} & \textbf{Value} \\
\hline
Device & Intel RealSense D405 \\
Product Line & D400 \\
Firmware Version & 5.17.0.10 \\
Depth Format & Z16 \\
IR Format & Y8 \\
Resolution & $848 \times 480$ \\
Frame Rate & 15 FPS \\
Schema Version & 1 \\
\hline
\end{tabular}
\end{table}

\begin{table}[H]
\centering
\caption{Intel RealSense D405 camera parameters.}
\label{tab:d405_parameters}
\scriptsize
\begin{tabular}{ll|ll}
\hline
\textbf{Parameter} & \textbf{Value} &
\textbf{Parameter} & \textbf{Value} \\
\hline
Auto Exposure & True &
Manual Exposure & 33000 \\
Depth Gain & 16 &
Auto White Balance & True \\
Depth Units & 100 &
Z Units & 100 \\
Depth Clamp Min & 0 &
Depth Clamp Max & 65536 \\
Disparity Mode & 0 &
Disparity Shift & 0 \\
Amplitude Factor & 0 &
Lambda AD & 800 \\
Lambda Census & 26 &
Left-Right Threshold & 24 \\
Median Threshold & 500 &
Neighbor Threshold & 7 \\
Min Score Threshold A & 1 &
Max Score Threshold B & 2047 \\
Second Peak Delta & 325 &
Texture Count Threshold & 0 \\
Texture Difference Threshold & 0 &
Use RSM & 1 \\
RAU Min E & 1 &
RAU Min N & 1 \\
RAU Min S & 1 &
RAU Min W & 1 \\
RAU Min NS Sum & 3 &
RAU Min WE Sum & 3 \\
Region Shrink U & 3 &
Region Shrink V & 1 \\
Region Color Threshold (R) & 0.0499022 &
Region Color Threshold (G) & 0.0499022 \\
Region Color Threshold (B) & 0.0499022 &
RSM Diff Threshold & 4 \\
RSM RAU SLO Diff Threshold & 1 &
RSM Remove Threshold & 0.375 \\
Scanline P1 & 60 &
Scanline P2 & 342 \\
Scanline P1 (1 Discon.) & 105 &
Scanline P2 (1 Discon.) & 190 \\
Scanline P1 (2 Discon.) & 70 &
Scanline P2 (2 Discon.) & 130 \\
Scanline Edge Tau (R) & 72 &
Scanline Edge Tau (G) & 72 \\
Scanline Edge Tau (B) & 72 &
Robbins--Monro Increment & 10 \\
Robbins--Monro Decrement & 10 &
Ignore SAD & 0 \\
\hline
\end{tabular}
\end{table}


% \bibliography{your_external_BibTeX_file}

%=====================================
% References, variant B: internal bibliography
%=====================================

% ACS format

\end{document}